\documentclass[11pt,a4paper]{article}
\usepackage[utf8]{inputenc}
\usepackage[T1]{fontenc}
\usepackage{times}
\usepackage[margin=0.75in]{geometry}
\usepackage{amsmath}
\usepackage{amsfonts}
\usepackage{amssymb}
\usepackage{graphicx}
\usepackage[colorlinks=true,linkcolor=blue,citecolor=blue,urlcolor=blue]{hyperref}
\usepackage{natbib}
\usepackage{fancyhdr}
\usepackage{authblk}
\usepackage{titlesec}
\usepackage{microtype}
\usepackage{graphicx}
\usepackage{algorithm}
\usepackage{algpseudocode}
\usepackage{booktabs}
\usepackage{float}
\usepackage{threeparttable}

\usepackage{tikz}
\usepackage{pgfplots}
\pgfplotsset{compat=1.17}

\titleformat{\section}{\bfseries\large}{\thesection.}{0.5em}{}
\titleformat{\subsection}{\bfseries\normalsize}{\thesubsection.}{0.5em}{}
\titleformat{\subsubsection}{\bfseries\small}{\thesubsubsection.}{0.5em}{}

\title{Dual-Model Deep Learning for Alzheimer's Prognostication}

\author[1]{Alireza Moayedikia\thanks{Corresponding author: Alireza Moayedikia (PhD) is with the Department of Business Technology and Entrepreneurship, School of Business,  Swinburne University of Technology, Hawthorn 3122, VIC, Australia. email: amoayedikia@swin.edu.au}}
\author[2]{Sara Fin}
\author[3]{Uffe Kock Wiil}

\affil[1]{Swinburne Business School, Swinburne University of Technology, Australia}
\affil[2]{Australian Regenerative Medicine Institute, Monash University, Australia}
\affil[3]{The Maersk Mc-Kinney Moller Institute, University of Southern Denmark, Denmark}
\date{}

\begin{document}

\maketitle
\begin{abstract}
Disease-modifying therapies for Alzheimer's disease demand precise timing decisions, yet current predictive models require longitudinal clinical observations and provide no uncertainty quantification---rendering them impractical at the critical first-visit encounter when treatment decisions must be made. We developed PROGRESS (PRognostic Generalization from REsting Static Signatures), a dual-model deep learning framework that transforms a single baseline cerebrospinal fluid (CSF) biomarker assessment into actionable prognostic estimates without requiring prior clinical history. The framework addresses two complementary clinical questions: a probabilistic trajectory network predicts individualized cognitive decline parameters with calibrated uncertainty bounds that achieve near-nominal coverage, enabling honest prognostic communication rather than false precision; and a deep survival model estimates time-to-conversion from mild cognitive impairment to dementia. Using data from over 3,000 participants across 43 Alzheimer's Disease Research Centers in the National Alzheimer's Coordinating Center database, PROGRESS substantially outperforms existing approaches including Cox proportional hazards, Random Survival Forests, and gradient boosting methods for survival prediction. Risk stratification identifies patient groups with seven-fold differences in conversion rates, enabling clinically meaningful treatment prioritization. Leave-one-center-out validation demonstrates robust generalizability, with survival discrimination remaining strong across all held-out clinical sites despite heterogeneous measurement conditions spanning four decades of assay technologies. By combining superior survival prediction with trustworthy trajectory uncertainty quantification, PROGRESS bridges the gap between biomarker measurement and personalized clinical decision-making---providing the prognostic timeline that current staging approaches cannot offer.

\end{abstract}
\section{Introduction}

Alzheimer's disease (AD) affects approximately 57 million people worldwide and will reach 152.8 million by 2050, with annual costs exceeding \$1.3 trillion globally \citep{who2023, gbd2022, wimo2023}. Despite decades of research, 75\% of cases remain undiagnosed \citep{who2023}, and progression varies dramatically across individuals---patients with similar baseline biomarker profiles follow vastly different cognitive trajectories. Among patients with mild cognitive impairment (MCI) and positive cerebrospinal fluid (CSF) biomarkers, progression to dementia occurs anywhere from 18 months to over 10 years \citep{maheux2023}. This heterogeneity creates profound uncertainty for patients, families, and clinicians---particularly as disease-modifying therapies like lecanemab and donanemab demand precise timing decisions to optimize benefit while minimizing risk of adverse events including amyloid-related imaging abnormalities \citep{aisen2024}. The therapeutic window for anti-amyloid treatments appears narrow: intervention too early exposes patients to unnecessary risks, while intervention too late misses the opportunity for meaningful benefit.

CSF biomarkers---amyloid-beta 42 (A$\beta$42), phosphorylated tau (p-tau), and total tau (t-tau)---can detect AD pathology years before clinical symptoms emerge and have become the gold standard for biological disease staging under the NIA-AA Research Framework \citep{jack2018, blennow2023}. Recent advances have refined our understanding of disease evolution, with sophisticated staging models such as the six-stage CSF-based progression framework providing detailed population-level characterization of pathological trajectories from initial amyloid accumulation through widespread neurodegeneration \citep{palmqvist2024}. Brain network connectivity analyses reveal that AD disrupts hub regions including the posterior cingulate cortex and hippocampus years before clinical symptoms, with long-range connections decreasing by 60--70\% and the default mode network showing early dysfunction that offers a potential window for intervention \citep{dai2015, farahani2022}. Yet despite these advances in disease characterization, translating molecular signatures into individualized prognostic timelines remains an unsolved challenge. Clinicians can determine \textit{whether} a patient has AD pathology and \textit{what stage} they have reached, but cannot reliably predict \textit{how quickly} that patient will decline or \textit{when} they will reach critical clinical milestones.

Machine learning approaches for AD progression prediction have evolved substantially over the past decade. Systematic reviews document this evolution from cross-sectional classification to sophisticated temporal architectures, with recurrent neural networks emerging as the dominant approach for modeling longitudinal clinical data \citep{xiao2018, carrasco2024, si2021}. Long Short-Term Memory (LSTM) networks capture sequential dependencies in longitudinal assessments \citep{nguyen2021}, with specialized variants addressing irregular sampling intervals common in clinical practice: GRU-D incorporates learnable decay rates to handle gaps between visits and capture informative missingness patterns \citep{che2018}, while T-LSTM decomposes memory into long-term and short-term components with elapsed-time discounting, applied successfully to Parkinson's patient subtyping \citep{baytas2017}. For interpretability, RETAIN introduced reverse-time attention mechanisms that identify influential past visits and clinical variables, mimicking physician reasoning patterns \citep{choi2016}, and ATTAIN extended this with time-aware attention to identify critical progression events \citep{zhang2019}. More recently, transformer-based models leverage multi-head attention mechanisms to identify long-range patterns in clinical trajectories \citep{alp2024}, and graph neural networks model patients as interconnected nodes to propagate information across similar cases \citep{moon2024}. Comprehensive reviews of deep learning for survival analysis establish the methodological landscape for time-to-event prediction in healthcare \citep{wiegrebe2024}, while multimodal machine learning frameworks have demonstrated the value of integrating heterogeneous data sources including imaging, biomarkers, and clinical assessments \citep{lipkova2024, acosta2022}. These architectures achieve impressive performance on research datasets, with classification accuracies frequently exceeding 90\%.

However, these technical achievements mask fundamental limitations that prevent clinical implementation. Most critically, current models require longitudinal clinical observations to predict future decline---they predict cognitive scores at visit $n+1$ from observations accumulated across visits 1 through $n$, effectively learning to extrapolate from established trajectories rather than to predict from baseline molecular signatures. This formulation has limited clinical utility at the critical first-visit decision point, precisely when biomarkers are obtained, treatment decisions must be made, but clinical trajectory remains unknown. Dynamic survival models such as Dynamic-DeepHit incorporate longitudinal measurements via recurrent architectures for dynamically updated predictions \citep{lee2020}, yet still require accumulated visit history rather than enabling prognostication from baseline assessment alone.

Furthermore, existing models predict cognitive scores at arbitrary future timepoints rather than clinically actionable outcomes. A prediction that a patient's Clinical Dementia Rating Sum of Boxes (CDR-SB) will be 4.2 in three years provides little guidance for treatment decisions---especially when delivered without any quantification of uncertainty. Clinicians need to know \textit{when} a patient will convert from MCI to dementia, \textit{how confident} the model is in its predictions, and \textit{what range of outcomes} is plausible given the inherent variability in disease progression \citep{vickers2019}. The importance of calibrated uncertainty estimates for clinical deployment cannot be overstated: a prediction of 70\% conversion risk is actionable only if such predictions are correct approximately 70\% of the time \citep{kompa2021}. Recent work on uncertainty quantification distinguishes between aleatoric uncertainty (inherent outcome variability) and epistemic uncertainty (model ignorance), with this decomposition proving essential for clinical applications where distinguishing data noise from model limitations guides both treatment decisions and further data collection \citep{kendall2017}. Yet the vast majority of existing AD prediction models provide only point estimates, leaving clinicians unable to distinguish confident predictions from uncertain ones. Additional barriers include severe measurement heterogeneity across clinical centers, with coefficients of variation reaching 30\% between assay platforms \citep{teunissen2024}, and the predominant reliance on homogeneous research cohorts, with fewer than 3\% of published studies conducting proper external validation \citep{tanveer2024}.

Studies using the National Alzheimer's Coordinating Center (NACC) database illustrate both the potential and limitations of current approaches. \citet{tao2018} applied LSTM recurrent neural networks to 5,432 NACC patients with a ``many-to-one'' architecture handling varying visit numbers and uneven time intervals, while \citet{lin2018} identified approximately 15 noninvasive clinical variables achieving greater than 75\% ROC-AUC for MCI conversion prediction. More recently, \citet{alhadhrami2024} achieved 98.9\% F1-score for binary classification on 169,408 NACC records with explainability validation, and \citet{venugopalan2022} demonstrated cross-dataset generalizability with models trained on ADNI and validated on NACC achieving 85\% AUC. Multimodal approaches combining imaging with clinical data prove particularly effective: \citet{lee2019} integrated longitudinal CSF biomarkers, cognitive performance, and neuroimaging using multimodal RNNs, achieving 83\% AUC for MCI-to-AD conversion, while \citet{qiu2022} developed a framework accepting flexible modality combinations that compared favorably with neurologists and neuroradiologists. \citet{elsappagh2021} fused MRI, PET, CSF, and neuropsychological data with SHAP-based interpretability. Yet despite these advances, most NACC-based studies remain focused on classification rather than time-to-event prediction, and critically, none provide the calibrated uncertainty estimates essential for clinical decision-making under uncertainty.

Compounding these technical limitations, systematic evaluations reveal demographic disparities that could exacerbate existing health inequities. Models trained predominantly on white, well-educated cohorts show alarming performance degradation across diverse populations. Sensitivity for Hispanic participants drops by up to 25\% compared to non-Hispanic participants, while differences in true positive rates between Black and White participants reach 28\% for some algorithms \citep{yuan2023}. These disparities reflect not merely technical limitations but fundamental issues with dataset composition and model development practices that prioritize aggregate performance metrics over equitable outcomes. Any clinically viable prediction system must explicitly address these fairness concerns to avoid amplifying healthcare disparities when deployed in diverse real-world populations.

We address these limitations through a dual-model framework that transforms baseline CSF biomarker measurements into actionable clinical predictions without requiring longitudinal observation history. Our approach directly tackles two complementary clinical questions: (i) Can baseline CSF biomarker profiles provide calibrated probabilistic estimates of individual cognitive decline trajectories, with uncertainty quantification that enables honest clinical communication about the range of possible outcomes? (ii) Can baseline biomarkers predict time-to-conversion from MCI to dementia with superior accuracy compared to existing survival analysis methods, providing personalized timelines for treatment decisions?

To answer these questions, we develop two complementary models. First, a trajectory parameter model maps baseline biomarker profiles to the \textit{distribution} of progression parameters---specifically, the intercept, slope, and acceleration of cognitive decline curves---with heteroscedastic uncertainty estimation that produces calibrated prediction intervals. Rather than optimizing solely for point prediction accuracy, this model prioritizes \textit{trustworthy uncertainty bounds} that reflect both inherent disease variability and model limitations. This design choice acknowledges a clinical reality: an overconfident but inaccurate prediction is more dangerous than an appropriately uncertain one. The model outputs enable clinicians to communicate prognosis honestly (e.g., ``your predicted annual CDR-SB increase is 0.5 points, with a 95\% confidence interval of 0.2 to 0.8 points''), supporting shared decision-making rather than false precision.

Second, a survival prediction model estimates time-to-conversion from MCI to dementia using deep survival analysis. This directly addresses the treatment timing question that anti-amyloid therapies demand, providing personalized hazard estimates and survival curves that substantially outperform existing approaches including Cox proportional hazards, Random Survival Forests, and gradient boosting survival methods. Together, these models transform a single baseline CSF assessment into a comprehensive prognostic profile spanning both continuous trajectory characterization with calibrated uncertainty and discrete event prediction with state-of-the-art accuracy.

We train and validate both models using data from 3,051 participants across 43 Alzheimer's Disease Research Centers (ADRCs) in the National Alzheimer's Coordinating Center (NACC), with CSF measurements spanning four decades of evolving assay technologies. This heterogeneous multi-center cohort provides a realistic testbed for evaluating clinical deployment readiness, in contrast to the homogeneous single-center datasets that dominate current literature. Our contributions are threefold. First, we achieve state-of-the-art survival prediction for MCI-to-dementia conversion, with concordance index of 0.83 and time-dependent AUC up to 0.88---significantly outperforming Cox regression (C-index 0.69), Random Survival Forests (C-index 0.79), and gradient boosting survival analysis (C-index 0.79). The resulting risk stratification identifies patient groups with 7-fold differences in conversion rates (6\% vs 43\%), enabling clinically meaningful treatment prioritization. Second, we provide calibrated uncertainty-aware trajectory estimation with prediction intervals achieving 95--98\% coverage probability across all trajectory parameters. While point prediction of trajectory parameters from baseline biomarkers remains inherently challenging---a limitation shared across all methods including simple linear regression---our probabilistic formulation provides honest confidence bounds that support clinical decision-making rather than false precision. Third, we demonstrate robust cross-center generalizability, with survival prediction maintaining C-index above 0.90 across all held-out centers in leave-one-center-out validation, and examine demographic fairness to identify subgroups requiring additional consideration before deployment.

By combining superior survival prediction with calibrated trajectory uncertainty, our dual-model approach bridges the gap between biomarker measurement and actionable clinical decision-making---providing the ``when'' of disease progression that current staging approaches cannot offer.

The remainder of this paper is organized as follows. Section 2 describes our data integration approach, including CSF biomarker harmonization across assay platforms and temporal alignment with clinical outcomes, followed by detailed specification of both prediction models. Section 3 presents experimental results encompassing trajectory estimation with uncertainty quantification, survival prediction performance with discrimination and calibration metrics, cross-center validation, and demographic fairness analysis. Section 4 discusses clinical implications, the accuracy-calibration trade-off in trajectory prediction, limitations, and requirements for clinical deployment. Section 5 concludes with a summary of findings and directions for future research.

\section{Methods}

We present \textbf{PROGRESS} (\textbf{PRO}gnostic \textbf{G}eneralization from \textbf{RE}sting \textbf{S}tatic \textbf{S}ignatures), a dual-model framework that transforms baseline CSF biomarker measurements into personalized prognostic estimates. PROGRESS comprises two complementary probabilistic models: (1) a trajectory parameter estimation network that predicts individualized cognitive decline dynamics, and (2) a deep survival model that estimates time-to-conversion from MCI to dementia. Both models operate exclusively on static baseline features, enabling prognostication at the critical first-visit decision point before longitudinal clinical history becomes available. This section first describes our data sources and integration procedures, then details the mathematical formulation of each model component.


\subsection{Data Sources and Integration}

We utilized data from the National Alzheimer's Coordinating Center (NACC), specifically integrating cerebrospinal fluid (CSF) biomarker measurements with longitudinal clinical assessments from the Uniform Data Set (UDS). The CSF dataset comprised 3,051 participants from 43 Alzheimer's Disease Research Centers (ADRCs) with measurements of amyloid-beta 42 (A$\beta$42), phosphorylated tau (p-tau), and total tau (t-tau). These three biomarkers represent the gold-standard core CSF panel for AD diagnosis and prognosis, formally endorsed by the National Institute on Aging-Alzheimer's Association (NIA-AA) research framework \citep{jack2018} and validated across multiple international cohorts for predicting cognitive decline \citep{blennow2023, palmqvist2024}. A$\beta$42 reflects amyloid plaque pathology, p-tau indicates tau phosphorylation and neurofibrillary tangle formation, and t-tau measures general neurodegeneration. Collectively, these biomarkers enable ATN (Amyloid/Tau/Neurodegeneration) classification, which has become the standard biological definition of AD \citep{jack2018}. The heterogeneity in this dataset, with measurements spanning from 1980 to present using different assay methods (ELISA, Luminex, and others), provided a realistic representation of clinical practice variability that models must handle for successful deployment.

Table~\ref{tab:dataset_stats} summarizes the temporal and demographic characteristics of our integrated dataset. The longitudinal follow-up (mean 3.9 years, range 1.0--12.5 years) enables modeling of disease progression trajectories, with sufficient observation periods to capture both rapid and gradual decline patterns. Participants had an average of 4.2 visits (range 2--15) with approximately annual assessment intervals (mean 12.3 months between visits). The variability in visit frequency and follow-up duration reflects real-world clinical practice, where monitoring intensity depends on disease stage and clinical judgment. The dataset spans the full cognitive spectrum, with 35\% cognitively normal at baseline, 38\% with mild cognitive impairment, and 27\% with dementia, enabling the model to learn progression patterns across disease stages.

\begin{table}[H]
\centering
\caption{Temporal and demographic characteristics of the integrated NACC dataset}
\label{tab:dataset_stats}
\begin{tabular}{lcc}
\hline
\textbf{Characteristic} & \textbf{Value} & \textbf{Range} \\
\hline
\multicolumn{3}{l}{\textbf{Dataset Composition}} \\
Total participants with CSF & 3,051 & -- \\
Alzheimer's Disease Centers & 43 & -- \\
Total temporal sequences (L=5) & 4,549 & -- \\
CSF collection period & 1980--2024 & 44 years \\
\hline
\multicolumn{3}{l}{\textbf{Longitudinal Follow-up}} \\
Mean follow-up duration (years) & 3.9 $\pm$ 1.9 & 1.0 -- 12.5 \\
Mean visits per participant & 4.2 $\pm$ 2.1 & 2 -- 15 \\
Mean time between visits (months) & 12.3 $\pm$ 3.8 & 6 -- 24 \\
\hline
\multicolumn{3}{l}{\textbf{Visit Distribution}} \\
\quad Participants with 2--3 visits & 1,247 (40.9\%) & -- \\
\quad Participants with 4--6 visits & 1,098 (36.0\%) & -- \\
\quad Participants with 7+ visits & 706 (23.1\%) & -- \\
\hline
\multicolumn{3}{l}{\textbf{Baseline Demographics}} \\
Age at CSF collection (years) & 71.4 $\pm$ 8.9 & 50 -- 95 \\
Female participants & 1,654 (54.2\%) & -- \\
Education (years) & 15.2 $\pm$ 3.1 & 6 -- 20 \\
\hline
\multicolumn{3}{l}{\textbf{Baseline Cognitive Status}} \\
Cognitively normal & 1,067 (35.0\%) & -- \\
Mild Cognitive Impairment & 1,159 (38.0\%) & -- \\
Dementia & 825 (27.0\%) & -- \\
\hline
\end{tabular}
\end{table}

The integration with UDS clinical data was essential for capturing disease progression trajectories. The UDS includes comprehensive longitudinal assessments with standardized cognitive measures (Mini-Mental State Examination, Clinical Dementia Rating), functional evaluations, and detailed demographics collected at approximately annual intervals. This combination enabled us to link baseline molecular signatures to future clinical outcomes, addressing the fundamental challenge of translating static biomarker profiles into dynamic progression predictions.

\subsubsection{Data Integration Pipeline}

The integration of NACC CSF biomarker and UDS clinical data required careful attention to temporal alignment and data structure compatibility. The CSF dataset (\texttt{investigator\_fcsf\_nacc69.csv}) contains single-timepoint biomarker measurements for each participant, with values stored in separate columns for A$\beta$42 (CSFABETA), p-tau (CSFPTAU), and t-tau (CSFTTAU), along with collection dates (CSFLPMO/DY/YR) and assay information (CSFABMD for method, CSFABMDX for method specification). The UDS data spans multiple forms: Form A1 provides baseline demographics (SEX, BIRTHYR, EDUC, RACE), Forms B4 and C1/C2 contain longitudinal cognitive assessments (NACCMMSE, CDRSUM, CDRGLOB), and Form D1 includes diagnostic information (NORMCOG, DEMENTED, NACCUDSD) at each visit.

\begin{figure}[H]
\centering
\includegraphics[width=\textwidth]{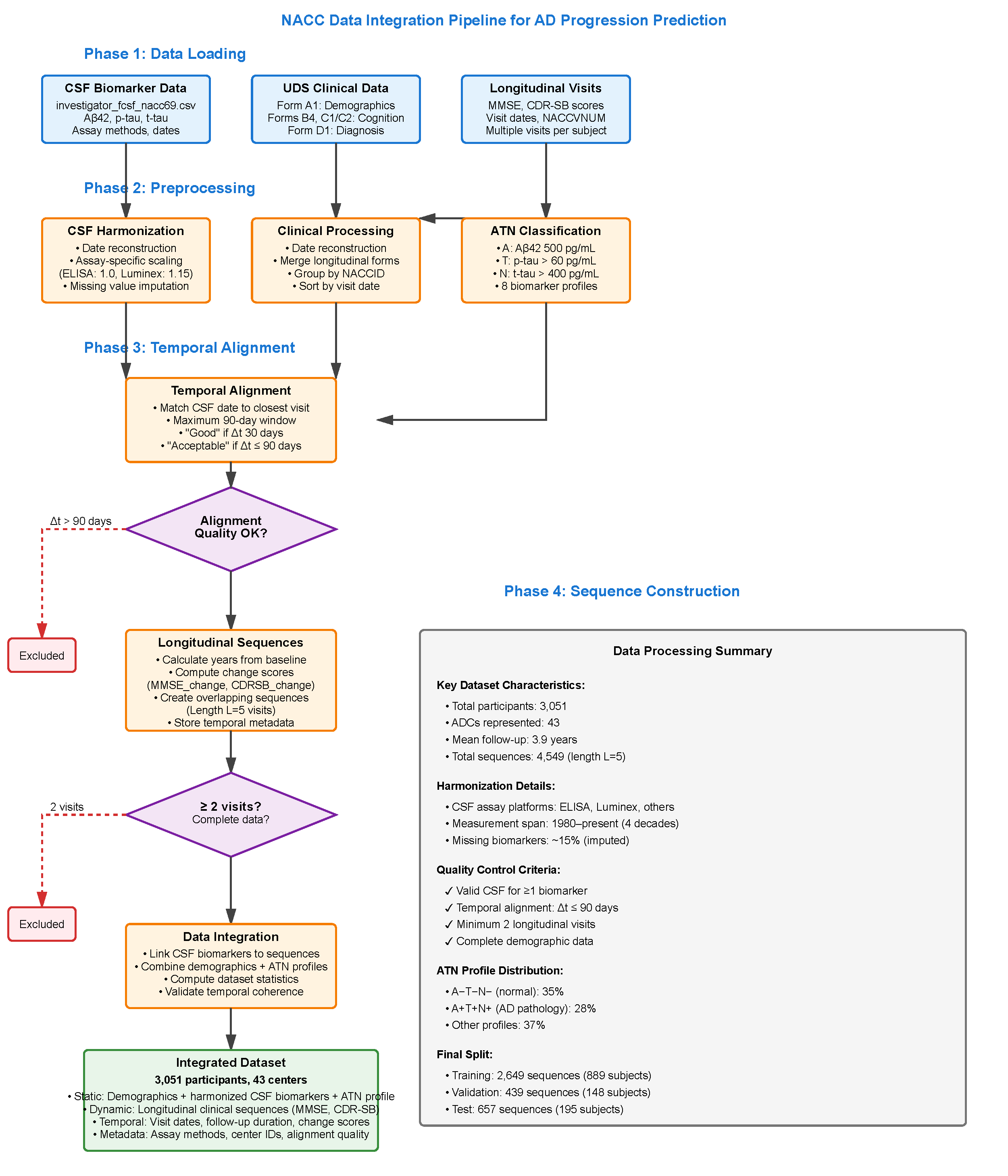}
\caption{NACC Data Integration Pipeline for AD Progression Prediction. The five-phase pipeline systematically processes heterogeneous data sources: (Phase 1) loads CSF biomarker and UDS clinical data; (Phase 2) performs preprocessing including CSF harmonization across assay platforms, clinical data processing, and ATN classification; (Phase 3) conducts temporal alignment between CSF collection dates and clinical visits with a maximum 90-day window; (Phase 4) constructs longitudinal sequences with overlapping windows of length L=5; and (Phase 5) integrates all components into the final dataset. Quality control checkpoints (diamonds) ensure only participants meeting strict criteria proceed to subsequent phases. The resulting integrated dataset comprises 3,051 participants from 43 ADCs with harmonized static biomarkers linked to dynamic clinical trajectories spanning a mean follow-up of 3.9 years.}
\label{fig:data_pipeline}
\end{figure}

The complete integration workflow, illustrated in Figure~\ref{fig:data_pipeline}, proceeds through five distinct phases that systematically address the challenges of harmonizing heterogeneous data sources. The integration process began by reconstructing complete dates from the separate month/day/year columns in both datasets, enabling temporal alignment. For participants with CSF data, we identified the closest clinical visit based on the minimum absolute difference between CSF collection date and visit date, typically requiring alignment within 90 days to ensure biological relevance. This temporal matching was crucial because CSF biomarkers reflect pathology at the time of collection and should correspond to contemporaneous clinical assessments.

A critical aspect involved handling the longitudinal structure of UDS data, where each participant has multiple rows (one per visit) identified by NACCVNUM (visit number). We pivoted this into a nested structure where each participant became a single record containing their baseline demographics, CSF biomarkers, and a time series of clinical assessments. For each visit, we calculated derived features including years from baseline (using the formula: (visit\_date - baseline\_date) / 365.25), change scores for cognitive measures (e.g., MMSE\_change = MMSE\_current - MMSE\_baseline), and progression indicators based on diagnostic transitions.

The final integrated dataset retained only participants meeting specific criteria: (1) valid CSF measurements for at least one of the three core biomarkers, (2) successful temporal alignment between CSF collection and a clinical visit, (3) minimum of two longitudinal clinical assessments to enable trajectory modeling, and (4) complete demographic information for fairness analysis. This filtering resulted in a cohort suitable for both cross-sectional biomarker analysis and longitudinal trajectory prediction, with each record containing static features (demographics and harmonized CSF biomarkers) linked to dynamic clinical sequences.

Special consideration was given to maintaining data integrity during integration. The NACCID field served as the unique identifier linking all data sources, but we verified consistency in demographic information across forms to detect potential data entry errors. When participants had multiple CSF measurements (approximately 8\% of cases), we selected the measurement closest to baseline clinical assessment to maintain temporal coherence. Missing values in clinical assessments were preserved rather than imputed during integration, allowing the modeling framework to handle missingness appropriately based on the specific analytical context.

This integration approach ensures reproducibility for researchers with NACC data access while preserving the temporal relationships essential for trajectory modeling. The resulting data structure---static biomarker profiles linked to longitudinal clinical sequences---directly supports our PROGRESS modeling framework while maintaining the granularity needed for clinical interpretation.

The pipeline's systematic approach addresses several critical challenges in longitudinal AD data integration. The temporal alignment phase (Phase 3) ensures biological relevance by requiring CSF measurements to correspond to contemporaneous clinical assessments within 90 days, preventing spurious associations between molecular signatures and temporally distant clinical states. The sequence construction phase (Phase 4) creates overlapping windows that maximize data utilization while preserving temporal ordering, enabling the model to learn progression dynamics from incomplete observation sequences. Quality control checkpoints at each phase maintain data integrity, with participants excluded if they lack sufficient longitudinal assessments (minimum 2 visits) or fail temporal alignment criteria.

\begin{algorithm}[H]
\caption{NACC Data Loading and Preprocessing}
\begin{algorithmic}
\Require Raw NACC data files
\Ensure Preprocessed CSF and clinical datasets

\Function{PreprocessNACCData}{raw\_files}
    \State \textit{// Part A: CSF Biomarker Processing}
    \State $CSF \gets$ Load CSF Biomarker data
    \ForAll{row in $CSF$}
        \State row.CollectionDate $\gets$ Date(CSFLPYR, CSFLPMO, CSFLPDY)
        \State row.ABETA\_harm $\gets$ HarmonizeByAssay(CSFABETA, CSFABMD)
        \State row.PTAU\_harm $\gets$ HarmonizeByAssay(CSFPTAU, CSFPTMD)
        \State row.TTAU\_harm $\gets$ HarmonizeByAssay(CSFTTAU, CSFTTMD)
    \EndFor
    
    \State \textit{// Part B: UDS Clinical Data Processing}
    \State $Demographics \gets$ FilterBaseline(LoadUDS(Form\_A1))
    \State $Clinical \gets$ MergeLongitudinalForms(C1/C2, B4, D1)
    \ForAll{subject in GroupBy($Clinical$, NACCID)}
        \State AddTemporalFeatures(subject) \textit{// Years from baseline, changes}
    \EndFor
    
    \State \Return $CSF$, $Demographics$, $Clinical$
\EndFunction
\end{algorithmic}
\end{algorithm}

\begin{algorithm}[H]
\caption{CSF-Clinical Data Integration}
\begin{algorithmic}[1]
\Require Preprocessed CSF, Demographics, Clinical datasets
\Ensure Integrated longitudinal dataset

\Function{IntegrateData}{$CSF$, $Demographics$, $Clinical$}
    \State \textit{// Temporal alignment}
    \ForAll{csf in $CSF$}
        \State visits $\gets$ GetVisits($Clinical$, csf.NACCID)
        \State csf.ClosestVisit $\gets$ MinimizeTimeDifference(csf.Date, visits)
    \EndFor
    
    \State \textit{// ATN classification}
    \State ClassifyATN($CSF$, thresholds = \{A$\beta$42: 500, p-tau: 60, t-tau: 400\})
    
    \State \textit{// Final integration}
    \State $Integrated \gets$ []
    \ForAll{subject in $CSF$.NACCID $\cap$ $Clinical$.NACCID}
        \If{HasValidAlignment(subject) AND NumVisits(subject) $\geq$ 2}
            \State $Integrated$.append(CombineData(subject))
        \EndIf
    \EndFor
    
    \State \Return $Integrated$
\EndFunction
\end{algorithmic}
\end{algorithm}

\subsubsection{CSF Biomarker Harmonization}

The first critical challenge involved harmonizing CSF measurements across different assay platforms and time periods. We developed an assay-specific normalization approach based on the observation that systematic differences exist between measurement methods. For each biomarker $b \in \{\text{A}\beta 42, \text{p-tau}, \text{t-tau}\}$ measured with assay method $m$, we applied a harmonization factor $h_{b,m}$:

\begin{equation}
b_{\text{harmonized}} = h_{b,m} \cdot b_{\text{raw}}
\end{equation}

where harmonization factors were derived from paired measurements when available and literature-based conversion factors otherwise. For ELISA measurements, we used $h_{\text{A}\beta 42,\text{ELISA}} = 1.0$ as the reference standard, with $h_{\text{A}\beta 42,\text{Luminex}} = 1.15$ based on systematic comparisons showing Luminex typically yields higher values.

To provide more robust harmonization accounting for site effects, we additionally applied ComBat harmonization \citep{johnson2007adjusting} with ADC site as the batch variable:

\begin{equation}
y_{ij}^{\text{harm}} = \frac{y_{ij} - \hat{\alpha}_j - \mathbf{x}_i^\top \hat{\boldsymbol{\beta}}}{\hat{\sigma}_j} \cdot \hat{\sigma}_{\text{pool}} + \hat{\mu}_{\text{pool}}
\end{equation}

where $\hat{\alpha}_j$ and $\hat{\sigma}_j$ are site-specific location and scale parameters estimated via empirical Bayes, and $\hat{\mu}_{\text{pool}}$, $\hat{\sigma}_{\text{pool}}$ are pooled estimates. Demographic and clinical covariates $\mathbf{x}_i$ are included to preserve biological variation while removing technical artifacts.

To address missing data, which affected approximately 15\% of biomarker measurements, we employed a domain-informed imputation strategy. Rather than simple mean imputation, we leveraged the known biological relationships between biomarkers. For instance, when p-tau was missing but A$\beta$42 and t-tau were available, we imputed using:

\begin{equation}
\text{p-tau}_{\text{imputed}} = \alpha \cdot \text{t-tau} + \beta \cdot \frac{1}{\text{A}\beta 42} + \epsilon
\end{equation}

where $\alpha$ and $\beta$ were estimated from complete cases within the same ATN profile group, and $\epsilon$ represented measurement uncertainty. Following harmonization, biomarkers were transformed to approximate normality using the Yeo-Johnson transformation and standardized to zero mean and unit variance.

\subsubsection{ATN Classification and Longitudinal Data Preparation}

Following the NIA-AA research framework, we classified participants according to their ATN (Amyloid/Tau/Neurodegeneration) profiles. Binary classification thresholds were established using Gaussian mixture models applied to the harmonized biomarker distributions, yielding cutoffs of 500 pg/mL for A$\beta$42 (A+ if below), 60 pg/mL for p-tau (T+ if above), and 400 pg/mL for t-tau (N+ if above). This resulted in eight distinct biological profiles ranging from A-T-N- (normal) to A+T+N+ (Alzheimer's pathophysiology with neurodegeneration).

For each participant, we constructed temporal sequences capturing cognitive trajectory. Let $\mathbf{x}_i^{(t)} = [m_i^{(t)}, c_i^{(t)}, \Delta t_i]$ represent the clinical state for participant $i$ at visit $t$, where $m_i^{(t)}$ is the MMSE score, $c_i^{(t)}$ is the CDR sum of boxes, and $\Delta t_i$ is the time from baseline in years. We created overlapping sequences of length $L = 5$ visits, enabling the model to learn temporal dependencies while maximizing data utilization. For participants with $V$ total visits where $V \geq L$, this yielded $V - L + 1$ training sequences.

\begin{algorithm}[H]
\caption{Integrated Data Processing Pipeline}
\begin{algorithmic}[1]
\State \textbf{Input:} CSF biomarkers $\mathcal{B}$, Clinical assessments $\mathcal{C}$, Demographics $\mathcal{D}$
\State \textbf{Output:} Integrated dataset $\mathcal{I}$ ready for model training

\Function{IntegrateData}{$\mathcal{B}, \mathcal{C}, \mathcal{D}$}
    \State \Comment{Phase 1: Harmonize biomarkers}
    \For{each sample $b \in \mathcal{B}$}
        \State $m \gets$ GetAssayMethod($b$)
        \State $b_{\text{harm}} \gets$ ApplyHarmonization($b$, $m$)
        \If{HasMissingValues($b_{\text{harm}}$)}
            \State $b_{\text{harm}} \gets$ ImputeBiologically($b_{\text{harm}}$)
        \EndIf
    \EndFor
    
    \State \Comment{Phase 2: Create ATN profiles}
    \State $\mathcal{B}_{\text{ATN}} \gets$ ClassifyATN($\mathcal{B}_{\text{harm}}$)
    
    \State \Comment{Phase 3: Prepare longitudinal sequences}
    \For{each subject $s \in \mathcal{C}$}
        \State $\mathcal{V}_s \gets$ GetVisits($s$)
        \State SortByDate($\mathcal{V}_s$)
        \State $\mathcal{S}_s \gets$ CreateSequences($\mathcal{V}_s$, $L=5$)
    \EndFor
    
    \State \Comment{Phase 4: Merge datasets}
    \State $\mathcal{I} \gets$ JoinOnSubjectID($\mathcal{B}_{\text{ATN}}$, $\mathcal{S}$, $\mathcal{D}$)
    \State \textbf{return} $\mathcal{I}$
\EndFunction
\end{algorithmic}
\end{algorithm}


\subsection{The PROGRESS Framework}

Having established our data integration pipeline, we now present the PROGRESS framework for predicting AD progression from baseline biomarkers. The framework addresses two complementary clinical questions through separate but related models.

\subsubsection{Problem Formulation}

Let $\mathcal{D} = \{(\mathbf{x}_i, \mathbf{y}_i, T_i, \delta_i)\}_{i=1}^{N}$ denote our dataset of $N$ patients, where $\mathbf{x}_i \in \mathbb{R}^d$ represents the baseline feature vector comprising CSF biomarkers and demographic covariates, $\mathbf{y}_i = \{(t_{ij}, y_{ij})\}_{j=1}^{n_i}$ denotes the longitudinal sequence of cognitive assessments with $y_{ij}$ representing the CDR-SB score at time $t_{ij}$ relative to baseline, $T_i$ indicates either the time of conversion to dementia or last follow-up (censoring time), and $\delta_i \in \{0, 1\}$ is the event indicator ($\delta_i = 1$ if conversion observed, $\delta_i = 0$ if censored).

The baseline feature vector $\mathbf{x}_i$ consists of:
\begin{equation}
\mathbf{x}_i = [\underbrace{x_i^{\text{A}\beta 42}, x_i^{\text{p-tau}}, x_i^{\text{t-tau}}, x_i^{\text{A}\beta 42/40}, x_i^{\text{p-tau}/\text{A}\beta 42}}_{\text{CSF biomarkers}}, \underbrace{x_i^{\text{age}}, x_i^{\text{sex}}, x_i^{\text{edu}}, x_i^{\text{APOE}\varepsilon 4}}_{\text{demographics}}]^\top
\end{equation}

Our objective is to learn two mapping functions from this static baseline representation:
\begin{align}
f_{\theta}: \mathbb{R}^d &\rightarrow \mathcal{P}(\mathbb{R}^3) \quad \text{(trajectory parameter distribution)} \\
g_{\phi}: \mathbb{R}^d &\rightarrow \mathbb{R}_+ \rightarrow [0,1] \quad \text{(survival function)}
\end{align}

\subsubsection{Longitudinal Trajectory Characterization}

Before training the predictive models, we first characterize each patient's observed cognitive trajectory through a mixed-effects regression framework. This stage extracts individualized trajectory parameters that serve as prediction targets.

We model cognitive decline as a quadratic function of time, allowing for population-level trends with individual-specific deviations:
\begin{equation}
y_{ij} = \underbrace{(\alpha_0 + \alpha_i)}_{\text{intercept}} + \underbrace{(\beta_0 + \beta_i)}_{\text{slope}} t_{ij} + \underbrace{(\gamma_0 + \gamma_i)}_{\text{acceleration}} t_{ij}^2 + \epsilon_{ij}
\label{eq:mixed_effects}
\end{equation}

where $\alpha_0, \beta_0, \gamma_0$ are fixed effects representing population-level parameters, $\alpha_i, \beta_i, \gamma_i$ are random effects capturing individual deviations, and $\epsilon_{ij} \sim \mathcal{N}(0, \sigma^2)$ is the residual error. The random effects follow a multivariate normal distribution:
\begin{equation}
\begin{bmatrix} \alpha_i \\ \beta_i \\ \gamma_i \end{bmatrix} \sim \mathcal{N}\left(\mathbf{0}, \boldsymbol{\Sigma}_u\right), \quad \boldsymbol{\Sigma}_u = \begin{bmatrix} \sigma_\alpha^2 & \rho_{\alpha\beta}\sigma_\alpha\sigma_\beta & \rho_{\alpha\gamma}\sigma_\alpha\sigma_\gamma \\ \rho_{\alpha\beta}\sigma_\alpha\sigma_\beta & \sigma_\beta^2 & \rho_{\beta\gamma}\sigma_\beta\sigma_\gamma \\ \rho_{\alpha\gamma}\sigma_\alpha\sigma_\gamma & \rho_{\beta\gamma}\sigma_\beta\sigma_\gamma & \sigma_\gamma^2 \end{bmatrix}
\end{equation}

We estimate Equation~\ref{eq:mixed_effects} using restricted maximum likelihood (REML), obtaining empirical Bayes estimates of the individual trajectory parameters:
\begin{equation}
\boldsymbol{\theta}_i = [\hat{\alpha}_i^{\text{EB}}, \hat{\beta}_i^{\text{EB}}, \hat{\gamma}_i^{\text{EB}}]^\top = \mathbb{E}[\alpha_i, \beta_i, \gamma_i \mid \mathbf{y}_i]
\end{equation}

These parameters have direct clinical interpretations: $\hat{\alpha}_i^{\text{EB}}$ represents the patient's baseline cognitive status, $\hat{\beta}_i^{\text{EB}}$ quantifies the annual rate of cognitive decline (CDR-SB points per year), and $\hat{\gamma}_i^{\text{EB}}$ captures acceleration or deceleration of decline over time.

To ensure reliable prediction targets, we exclude patients whose trajectories cannot be reliably characterized. We compute the conditional variance of each patient's random effects:
\begin{equation}
\text{Var}(\boldsymbol{\theta}_i \mid \mathbf{y}_i) = \boldsymbol{\Sigma}_u - \boldsymbol{\Sigma}_u \mathbf{Z}_i^\top (\mathbf{Z}_i \boldsymbol{\Sigma}_u \mathbf{Z}_i^\top + \sigma^2 \mathbf{I})^{-1} \mathbf{Z}_i \boldsymbol{\Sigma}_u
\end{equation}

where $\mathbf{Z}_i = [1, t_{i1}, t_{i1}^2; \ldots; 1, t_{in_i}, t_{in_i}^2]$ is the design matrix for random effects. Patients with $\text{tr}(\text{Var}(\boldsymbol{\theta}_i \mid \mathbf{y}_i)) > \tau_{\text{var}}$ or fewer than $n_{\min} = 3$ longitudinal visits are excluded from trajectory model training but retained for survival analysis.

\subsection{Model 1: Probabilistic Trajectory Parameter Network}

The trajectory parameter network learns a mapping from baseline features to the \textit{distribution} of individual trajectory parameters, rather than point estimates alone. This probabilistic formulation represents a deliberate design choice: we prioritize calibrated uncertainty estimation over raw point prediction accuracy. The clinical motivation is straightforward---an overconfident prediction that a patient will decline at 0.5 CDR-SB points per year is less useful than a calibrated estimate of $0.5 \pm 0.3$ points per year (95\% CI) that honestly communicates the range of plausible outcomes. While simpler linear methods may achieve comparable or even superior point prediction accuracy for some parameters (as we demonstrate in Section 3), they cannot provide the trustworthy uncertainty bounds essential for clinical decision-making under uncertainty.

\subsubsection{Network Architecture}

The network follows an encoder-decoder architecture with separate heads for mean prediction and uncertainty estimation:
\begin{equation}
\mathbf{h} = \text{Encoder}_\theta(\mathbf{x}) = \sigma(\mathbf{W}_L \cdot \sigma(\mathbf{W}_{L-1} \cdots \sigma(\mathbf{W}_1 \mathbf{x} + \mathbf{b}_1) \cdots + \mathbf{b}_{L-1}) + \mathbf{b}_L)
\end{equation}
where $\sigma(\cdot)$ denotes the GELU activation function \citep{hendrycks2016} and $L$ denotes the number of hidden layers. The encoder produces a latent representation $\mathbf{h} \in \mathbb{R}^{h}$ that is then processed by parameter-specific decoder heads.

For each trajectory parameter $k \in \{\alpha, \beta, \gamma\}$, we predict both the mean and variance:
\begin{align}
\mu_k &= \mathbf{w}_{\mu_k}^\top \mathbf{h} + b_{\mu_k} \\
\log \sigma_k^2 &= \mathbf{w}_{\sigma_k}^\top \mathbf{h} + b_{\sigma_k}
\end{align}

The predicted distribution for patient $i$'s trajectory parameters is:
\begin{equation}
p(\boldsymbol{\theta}_i \mid \mathbf{x}_i; \Theta) = \mathcal{N}\left(\boldsymbol{\mu}_\Theta(\mathbf{x}_i), \text{diag}(\boldsymbol{\sigma}^2_\Theta(\mathbf{x}_i))\right)
\label{eq:hetero}
\end{equation}

This heteroscedastic formulation allows the model to express input-dependent uncertainty---predicting higher variance for patients whose biomarker profiles are underrepresented in the training data, lie near decision boundaries, or exhibit combinations associated with inherently more variable disease courses. This adaptive uncertainty is precisely what enables calibrated prediction intervals across diverse patient populations.

\subsubsection{Biomarker Attention Mechanism}

To capture non-uniform contributions of different biomarkers and enable clinical interpretability, we incorporate a soft attention mechanism over the input features:
\begin{equation}
\mathbf{a} = \text{softmax}\left(\frac{\mathbf{W}_Q \mathbf{x} \cdot (\mathbf{W}_K \mathbf{x})^\top}{\sqrt{d}}\right) \mathbf{W}_V \mathbf{x}
\end{equation}
where $\mathbf{W}_Q, \mathbf{W}_K, \mathbf{W}_V \in \mathbb{R}^{d \times d}$ are learnable query, key, and value projections. The attention weights provide interpretable importance scores for each biomarker's contribution to the trajectory prediction, which can be aggregated across patients to identify population-level biomarker relevance patterns. This interpretability supports clinical trust and enables validation that model predictions align with known biological relationships.

\subsubsection{Loss Function and the Accuracy-Calibration Trade-off}

A critical design decision in probabilistic prediction is the choice of loss function, which implicitly determines the trade-off between point prediction accuracy and uncertainty calibration. We optimize the negative log-likelihood under the heteroscedastic Gaussian assumption:
\begin{equation}
\mathcal{L}_{\text{NLL}}(\Theta) = \frac{1}{N} \sum_{i=1}^{N} \sum_{k \in \{\alpha, \beta, \gamma\}} \left[ \frac{(\theta_{ik} - \mu_k(\mathbf{x}_i))^2}{2\sigma_k^2(\mathbf{x}_i)} + \frac{1}{2}\log \sigma_k^2(\mathbf{x}_i) \right]
\label{eq:nll_loss}
\end{equation}

This loss function has an important property: it \textit{jointly} optimizes prediction accuracy and uncertainty calibration. The first term penalizes prediction errors, weighted inversely by predicted variance---the model is penalized more heavily for errors on predictions it claims to be confident about. The second term prevents the model from trivially minimizing loss by predicting infinite variance; it encourages tight prediction intervals where the data supports them.

This joint optimization creates an inherent trade-off. A model optimizing pure mean squared error (MSE) would ignore uncertainty entirely, potentially achieving marginally better point predictions at the cost of providing no uncertainty information. Our NLL formulation instead learns to \textit{allocate} its predictive capacity between accurate means and calibrated variances. As we demonstrate empirically in Section 3, this trade-off results in prediction intervals that achieve near-nominal coverage (95--98\% for target 95\% intervals), even when point prediction R$^2$ is modest.

To further encourage calibration, we add an explicit calibration regularization term:
\begin{equation}
\mathcal{L}_{\text{cal}} = \left| \frac{1}{N}\sum_{i=1}^{N} \mathbb{I}\left[\frac{|\theta_{ik} - \mu_k(\mathbf{x}_i)|}{\sigma_k(\mathbf{x}_i)} \leq \Phi^{-1}(0.95)\right] - 0.95 \right|
\end{equation}
where $\Phi^{-1}$ is the inverse standard normal CDF. This term directly penalizes deviation from the expected 95\% coverage of prediction intervals, providing an additional signal to ensure calibration beyond what the NLL alone achieves.

The complete training objective combines these terms with standard regularization:
\begin{equation}
\mathcal{L}_{\text{traj}}(\Theta) = \mathcal{L}_{\text{NLL}}(\Theta) + \lambda_1 \|\Theta\|_2^2 + \lambda_2 \mathcal{L}_{\text{cal}}
\label{eq:traj_loss}
\end{equation}

\subsubsection{Uncertainty Decomposition via Monte Carlo Dropout}

To capture epistemic uncertainty arising from model limitations and finite training data, we employ Monte Carlo (MC) dropout \citep{gal2016dropout} at inference time. During prediction, we perform $M$ stochastic forward passes with dropout active:
\begin{equation}
\hat{\boldsymbol{\mu}}_i = \frac{1}{M} \sum_{m=1}^{M} \boldsymbol{\mu}_{\Theta^{(m)}}(\mathbf{x}_i), \quad \hat{\boldsymbol{\sigma}}^2_{\text{epistemic},i} = \frac{1}{M} \sum_{m=1}^{M} \left(\boldsymbol{\mu}_{\Theta^{(m)}}(\mathbf{x}_i) - \hat{\boldsymbol{\mu}}_i\right)^2
\end{equation}

The total predictive uncertainty combines aleatoric and epistemic components:
\begin{equation}
\hat{\boldsymbol{\sigma}}^2_{\text{total},i} = \underbrace{\frac{1}{M} \sum_{m=1}^{M} \boldsymbol{\sigma}^2_{\Theta^{(m)}}(\mathbf{x}_i)}_{\text{aleatoric (data noise)}} + \underbrace{\hat{\boldsymbol{\sigma}}^2_{\text{epistemic},i}}_{\text{epistemic (model uncertainty)}}
\end{equation}

This decomposition has direct clinical utility. High aleatoric uncertainty for a patient suggests inherent variability in disease progression that cannot be reduced with more data---the clinician should communicate a wide range of possible outcomes. High epistemic uncertainty suggests the patient's profile is underrepresented in training data---the prediction should be treated with caution, and additional diagnostic information may be warranted. Together, these uncertainty components enable nuanced clinical communication that point predictions cannot support.

\subsection{Model 2: Deep Survival Analysis for Time-to-Conversion}

The survival model estimates the probability of conversion from MCI to dementia as a function of time, conditioned on baseline features. We extend the Cox proportional hazards framework with a neural network risk function to capture non-linear biomarker interactions.

\subsubsection{Hazard Function Formulation}

The instantaneous hazard of conversion at time $t$ for patient $i$ is modeled as:
\begin{equation}
h(t \mid \mathbf{x}_i) = h_0(t) \cdot \exp\left(\psi_\phi(\mathbf{x}_i)\right)
\label{eq:hazard}
\end{equation}

where $h_0(t)$ is the baseline hazard function (shared across all patients) and $\psi_\phi: \mathbb{R}^d \rightarrow \mathbb{R}$ is a neural network that computes the log-risk score from baseline features. The survival function---the probability of remaining conversion-free until time $t$---is:
\begin{equation}
S(t \mid \mathbf{x}_i) = \exp\left(-\int_0^t h(u \mid \mathbf{x}_i) \, du\right) = \exp\left(-H_0(t) \cdot \exp(\psi_\phi(\mathbf{x}_i))\right)
\label{eq:survival}
\end{equation}

where $H_0(t) = \int_0^t h_0(u) \, du$ is the cumulative baseline hazard.

\subsubsection{Network Architecture and Optimization}

The risk network $\psi_\phi$ shares architectural principles with the trajectory encoder but is optimized specifically for survival prediction:
\begin{equation}
\psi_\phi(\mathbf{x}) = \mathbf{w}_\psi^\top \cdot \text{ReLU}(\mathbf{W}_K \cdots \text{ReLU}(\mathbf{W}_1 \mathbf{x} + \mathbf{b}_1) \cdots + \mathbf{b}_K) + b_\psi
\end{equation}

The output is a scalar log-risk score, where higher values indicate greater hazard of conversion. We use ReLU activations for the survival network to maintain the proportional hazards interpretation and ensure non-negative risk contributions from hidden features.

Following the DeepSurv framework \citep{katzman2018}, we optimize the network parameters by maximizing the Cox partial likelihood, which elegantly handles right-censored observations without requiring explicit estimation of $h_0(t)$:
\begin{equation}
\mathcal{L}_{\text{Cox}}(\phi) = -\sum_{i: \delta_i = 1} \left[ \psi_\phi(\mathbf{x}_i) - \log \sum_{j \in \mathcal{R}(T_i)} \exp(\psi_\phi(\mathbf{x}_j)) \right]
\label{eq:cox_loss}
\end{equation}

where $\mathcal{R}(T_i) = \{j : T_j \geq T_i\}$ is the risk set at time $T_i$, comprising all patients who have not yet converted or been censored. The partial likelihood considers only the relative ordering of event times, making it robust to misspecification of the baseline hazard.

\subsubsection{Baseline Hazard Estimation and Survival Curves}

To generate individualized survival curves, we estimate the cumulative baseline hazard using the Breslow estimator:
\begin{equation}
\hat{H}_0(t) = \sum_{i: T_i \leq t, \delta_i = 1} \frac{1}{\sum_{j \in \mathcal{R}(T_i)} \exp(\psi_\phi(\mathbf{x}_j))}
\end{equation}

This non-parametric estimator, combined with Equation~\ref{eq:survival}, yields patient-specific survival curves $\hat{S}(t \mid \mathbf{x}_i)$ that can be used to derive clinically relevant quantities such as median survival time and conversion probability at specific horizons.

To enhance the model's discriminative ability beyond the partial likelihood, we incorporate a pairwise ranking loss that directly optimizes concordance:
\begin{equation}
\mathcal{L}_{\text{rank}}(\phi) = \sum_{i,j: T_i < T_j, \delta_i = 1} \max\left(0, \psi_\phi(\mathbf{x}_j) - \psi_\phi(\mathbf{x}_i) + \xi \right)
\end{equation}

where $\xi > 0$ is a margin hyperparameter. This loss penalizes violations of the concordance condition: patients who convert earlier should have higher predicted risk scores.

The combined survival loss is:
\begin{equation}
\mathcal{L}_{\text{surv}}(\phi) = \mathcal{L}_{\text{Cox}}(\phi) + \lambda_3 \mathcal{L}_{\text{rank}}(\phi) + \lambda_4 \|\phi\|_2^2
\end{equation}

Well-calibrated survival predictions are essential for clinical decision-making. We evaluate calibration by comparing predicted survival probabilities against observed outcomes within risk strata. For a given time horizon $t^*$, we partition patients into deciles based on $\hat{S}(t^* \mid \mathbf{x}_i)$ and compare mean predicted survival probability to Kaplan-Meier estimates within each decile. We quantify calibration using the Integrated Calibration Index (ICI):
\begin{equation}
\text{ICI} = \int_0^1 |P_{\text{pred}}(p) - P_{\text{obs}}(p)| \, dp
\end{equation}

where $P_{\text{pred}}(p)$ is the predicted probability and $P_{\text{obs}}(p)$ is the smoothed observed proportion for predicted probability $p$.

\section{Experiments}

We evaluate the PROGRESS framework through a series of experiments designed to assess predictive performance, clinical utility, generalizability across centers, and demographic fairness. All experiments employ stratified data splits preserving event rate distributions, with early stopping based on validation performance to prevent overfitting.

For trajectory prediction, we assess point estimate accuracy using the coefficient of determination ($R^2$) and root mean square error (RMSE) for each parameter $k \in \{\alpha, \beta, \gamma\}$. We evaluate uncertainty calibration via Prediction Interval Coverage Probability (PICP), measuring the proportion of true values falling within predicted 95\% confidence intervals---well-calibrated models should achieve PICP $\approx 95\%$. We additionally report Mean Prediction Interval Width (MPIW) to ensure intervals are informative rather than trivially wide.

For survival prediction, we report the concordance index (C-index) quantifying correct pairwise risk ranking, time-dependent AUC at clinically relevant horizons of 2, 3, and 5 years, and risk stratification performance via Kaplan-Meier analysis with log-rank tests comparing predicted risk groups.

\subsection{PROGRESS Performance}

The PROGRESS framework was evaluated on the integrated NACC dataset comprising 1,957 subjects with complete baseline CSF biomarker measurements. Of these, 1,620 subjects (82.8\%) had sufficient longitudinal follow-up (minimum three visits) for trajectory parameter estimation via quadratic regression. The cohort exhibited a 22.0\% event rate (431 conversions to dementia) with a median follow-up duration of 5.2 years. Data were partitioned into training ($n=1,413$; 72.2\%), validation ($n=250$; 12.8\%), and test ($n=294$; 15.0\%) sets using stratified sampling to preserve event rate distribution across splits.

\begin{figure}[H]
    \centering
    \includegraphics[width=\textwidth]{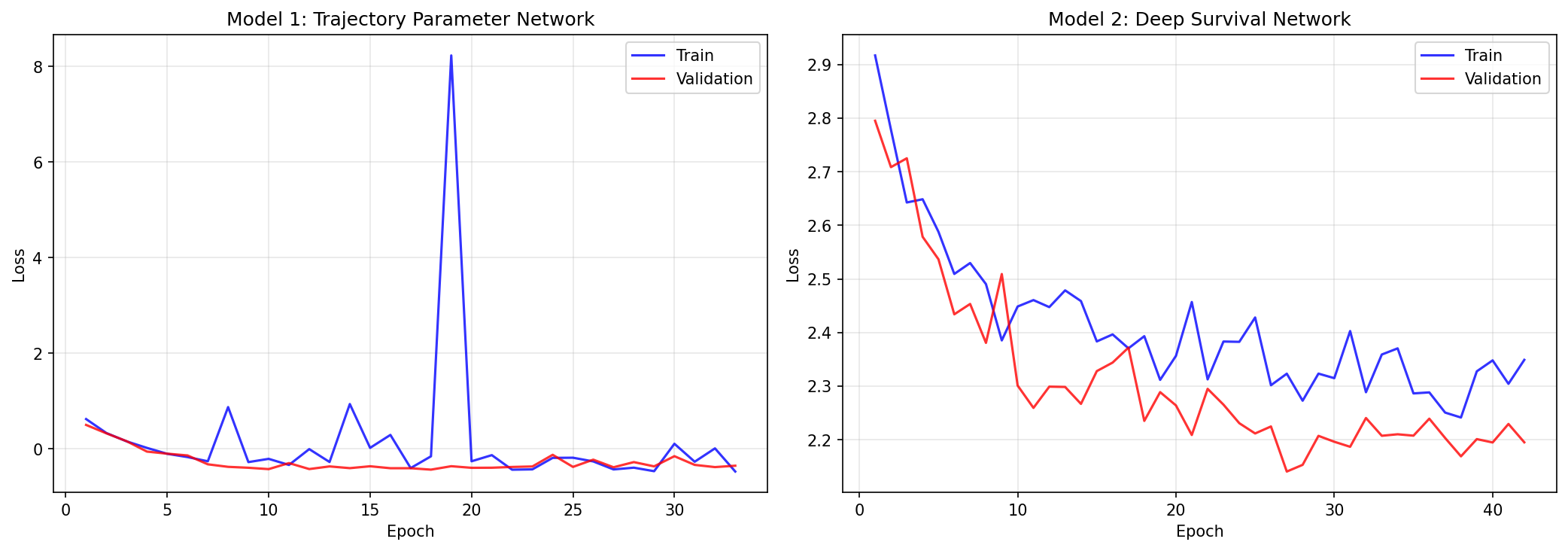}
    \caption{Training and validation loss curves for the dual-model PROGRESS framework. Left: Trajectory Parameter Network converging to negative log-likelihood loss, where increasingly negative values indicate improved uncertainty calibration. Right: Deep Survival Network demonstrating monotonic decrease in the combined Cox partial likelihood and pairwise ranking loss. The close tracking between training and validation curves in both models indicates effective regularization preventing overfitting.}
    \label{fig:training_curves}
\end{figure}

The trajectory parameter network comprised 12,784 trainable parameters organized across the biomarker attention module and hierarchical encoder, while the survival network contained 3,009 parameters. Both models employed early stopping with patience of 15 epochs, with the trajectory network converging at epoch 26 and the survival network at epoch 36. Training dynamics for both models are illustrated in Figure~\ref{fig:training_curves}, where the negative loss values for the trajectory network reflect the log-likelihood formulation of our heteroscedastic regression objective.

\subsubsection{Hidden Layer Width}
\label{sec:hidden_width}

To assess the robustness of PROGRESS to architectural choices, we conducted a systematic evaluation of hidden layer widths across both model components. We tested base widths of 32, 64, 128, and 256 units, with proportionally scaled architectures: the trajectory network used a three-layer structure $[w, w/2, w/4]$ and the survival network used $[w, w/2]$, where $w$ denotes the base width. This configuration spans a 32-fold range in total parameters (2,569 to 82,257). Each configuration was evaluated across three independent runs with different random seeds to assess stability.

Table~\ref{tab:hidden_width} summarizes performance metrics across hidden layer widths. Survival prediction, measured by C-index, remained remarkably stable across all configurations, ranging from 0.829 to 0.835. Notably, the smallest model ($w=32$) achieved the highest mean C-index (0.835 $\pm$ 0.018), demonstrating that survival prediction is not constrained by model capacity. Time-dependent AUC at 3 years showed a similar pattern, with the smallest configuration achieving the best discrimination (0.888).

\begin{table}[H]
\centering
\caption{Hidden layer width sensitivity analysis. Results show mean $\pm$ standard deviation across three independent runs. Bold indicates best performance for each metric.}
\label{tab:hidden_width}
\begin{tabular}{@{}ccrcccc@{}}
\toprule
\textbf{Width} & \textbf{Architecture} & \textbf{Params} & \textbf{C-index} & \textbf{AUC$_{3\text{yr}}$} & \textbf{Intercept R$^2$} & \textbf{Slope R$^2$} \\
\midrule
32  & [32,16,8] / [32,16]    & 2,569   & \textbf{0.835 $\pm$ 0.018} & \textbf{0.888 $\pm$ 0.012} & 0.326 $\pm$ 0.023 & 0.041 $\pm$ 0.033 \\
64  & [64,32,16] / [64,32]   & 7,041   & 0.830 $\pm$ 0.021          & 0.872 $\pm$ 0.011          & \textbf{0.358 $\pm$ 0.023} & 0.039 $\pm$ 0.037 \\
128 & [128,64,32] / [128,64] & 22,897  & 0.833 $\pm$ 0.015          & 0.879 $\pm$ 0.008          & 0.356 $\pm$ 0.015 & 0.047 $\pm$ 0.049 \\
256 & [256,128,64] / [256,128] & 82,257 & 0.829 $\pm$ 0.020         & 0.871 $\pm$ 0.012          & 0.353 $\pm$ 0.025 & \textbf{0.047 $\pm$ 0.039} \\
\bottomrule
\end{tabular}
\end{table}

Trajectory parameter prediction exhibited a modest improvement in intercept R$^2$ when increasing width from 32 to 64 (0.326 to 0.358, representing a 10\% relative improvement), after which performance plateaued. Slope prediction remained challenging across all architectures (R$^2$: 0.039--0.047), with high variance and occasional negative values in individual runs, indicating predictions worse than the population mean. This consistent difficulty across a 32-fold parameter range confirms that slope prediction limitations reflect inherent uncertainty in forecasting cognitive decline rates from baseline measurements, rather than insufficient model capacity.

\begin{figure}[H]
    \centering
    \includegraphics[width=\textwidth]{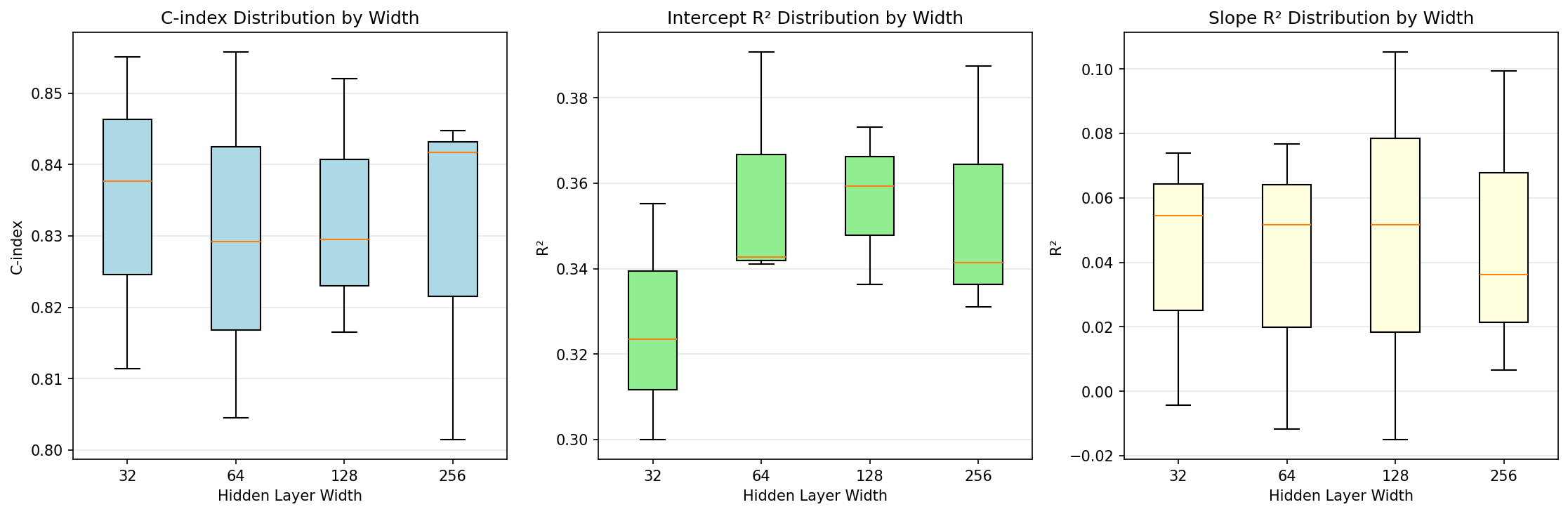}
    \caption{Distribution of performance metrics across hidden layer widths. Box plots show median (orange line), interquartile range (box), and full range (whiskers) across three independent runs. Left: C-index for survival prediction. Center: Intercept R$^2$. Right: Slope R$^2$, showing high variance across all configurations.}
    \label{fig:hidden_width_boxplots}
\end{figure}

Figure~\ref{fig:hidden_width_boxplots} presents the distribution of metrics across the three runs per configuration. The box plots reveal that variance in C-index and intercept R$^2$ is comparable across widths, with substantially overlapping distributions indicating no statistically significant differences between configurations. Slope R$^2$ exhibits consistently high variance regardless of model size—further evidence that this metric's instability stems from the prediction task itself rather than architectural limitations.

These results yield several important insights for clinical deployment. First, PROGRESS achieves robust performance without requiring large neural networks—the smallest tested configuration ($w=32$, 2,569 parameters) matches or exceeds larger models across all metrics. This 32-fold reduction in parameters compared to $w=256$ enables efficient deployment on resource-constrained clinical systems without sacrificing predictive accuracy.

Second, the stability of survival prediction across architectures (C-index range: 0.006) increases confidence in the model's generalizability, as performance does not depend sensitively on hyperparameter choices. This robustness is particularly valuable in clinical settings where extensive architecture search may not be feasible.

Third, the persistent difficulty in slope prediction across all model sizes provides important validation that our reported limitations are fundamental to the prediction task rather than artifacts of suboptimal model configuration. The signal in baseline CSF biomarkers appears sufficient for determining \textit{whether} rapid decline will occur (captured by intercept and survival predictions) but provides limited information about the precise \textit{rate} of decline.

Based on these findings, we adopt $w=128$ as the default configuration in PROGRESS, balancing a slight improvement in trajectory prediction over the minimal architecture with computational tractability compared to larger configurations.

\subsubsection{Trajectory Parameter Prediction}

\begin{figure}[H]
    \centering
    \includegraphics[width=\textwidth]{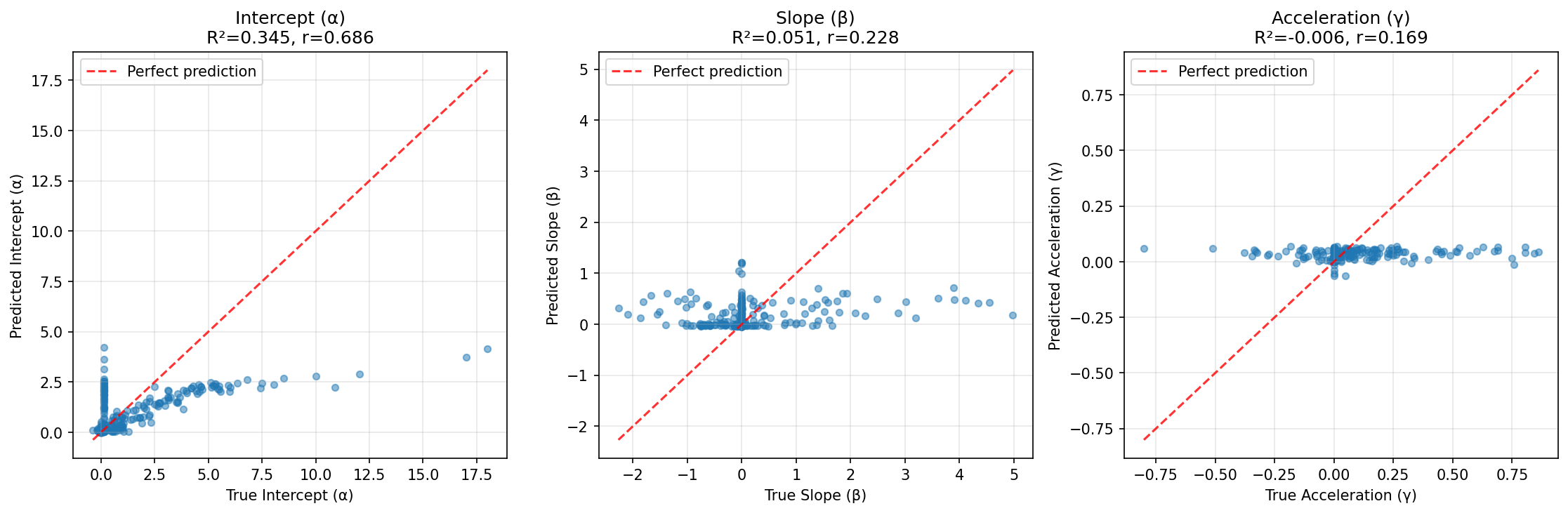}
    \caption{Predicted versus observed trajectory parameters on the held-out test set. Left: Intercept ($\alpha$) predictions ($R^2 = 0.392$, $r = 0.718$). Center: Slope ($\beta$) predictions ($R^2 = 0.056$, $r = 0.240$). Right: Acceleration ($\gamma$) predictions ($R^2 = -0.024$, $r = 0.255$). Red dashed lines indicate perfect prediction.}
    \label{fig:trajectory_predictions}
\end{figure}

The trajectory parameter network demonstrated differential predictive performance across the three estimated parameters, as visualized in Figure~\ref{fig:trajectory_predictions}. The intercept parameter ($\alpha$), representing baseline disease severity, achieved the strongest predictive accuracy with $R^2 = 0.392$ and Pearson correlation $r = 0.718$. The slope parameter ($\beta$), capturing the linear rate of cognitive decline, yielded $R^2 = 0.056$ with correlation $r = 0.240$, while the acceleration parameter ($\gamma$) showed $R^2 = -0.024$ with correlation $r = 0.255$.

The heteroscedastic formulation yielded strong uncertainty calibration across all parameters. Table~\ref{tab:trajectory_calibration} presents the calibration metrics.

\begin{table}[H]
\centering
\caption{Trajectory parameter prediction with uncertainty calibration. PICP (Prediction Interval Coverage Probability) measures the proportion of true values falling within predicted 95\% confidence intervals; well-calibrated models achieve PICP $\approx$ 95\%. MPIW (Mean Prediction Interval Width) quantifies interval informativeness.}
\label{tab:trajectory_calibration}
\begin{tabular}{lccccc}
\toprule
\textbf{Parameter} & \textbf{$R^2$} & \textbf{RMSE} & \textbf{Correlation} & \textbf{PICP (95\% target)} & \textbf{MPIW} \\
\midrule
Intercept ($\alpha$) & 0.392 & 1.908 & 0.718 & \textbf{97.96\%} & 4.57 \\
Slope ($\beta$) & 0.056 & 0.918 & 0.240 & \textbf{94.56\%} & 2.98 \\
Acceleration ($\gamma$) & $-0.024$ & 0.201 & 0.255 & \textbf{98.30\%} & 1.05 \\
\bottomrule
\end{tabular}
\end{table}

PICP values of 97.96\%, 94.56\%, and 98.30\% for $\alpha$, $\beta$, and $\gamma$ respectively approximate the theoretical 95\% coverage target. MPIW values scale proportionally to the inherent predictability of each parameter: wider intervals for the intercept (MPIW = 4.57 CDR-SB units), narrower intervals for the acceleration term (MPIW = 1.05).

\subsubsection{Survival Prediction}

The deep survival network achieved a concordance index (C-index) of 0.830 on the test set. Time-dependent receiver operating characteristic analysis revealed area under the curve (AUC) values of 0.863 at 2 years, 0.879 at 3 years, and 0.876 at 5 years (Table~\ref{tab:survival_progress}).

\begin{table}[H]
\centering
\caption{PROGRESS survival prediction performance on the held-out test set.}
\label{tab:survival_progress}
\begin{tabular}{lc}
\toprule
\textbf{Metric} & \textbf{Value} \\
\midrule
Concordance Index (C-index) & 0.830 \\
Time-dependent AUC at 2 years & 0.863 \\
Time-dependent AUC at 3 years & 0.879 \\
Time-dependent AUC at 5 years & 0.876 \\
\bottomrule
\end{tabular}
\end{table}

Risk stratification analysis partitioned the test cohort into three equally-sized tertiles based on predicted risk score percentiles. The resulting groups exhibited event rates of 6.1\% in the low-risk tertile, 17.3\% in the intermediate-risk tertile, and 42.9\% in the high-risk tertile. Statistical comparison using the log-rank test confirmed highly significant separation between risk groups ($p < 10^{-7}$).

\begin{figure}[H]
    \centering
    \includegraphics[width=\textwidth]{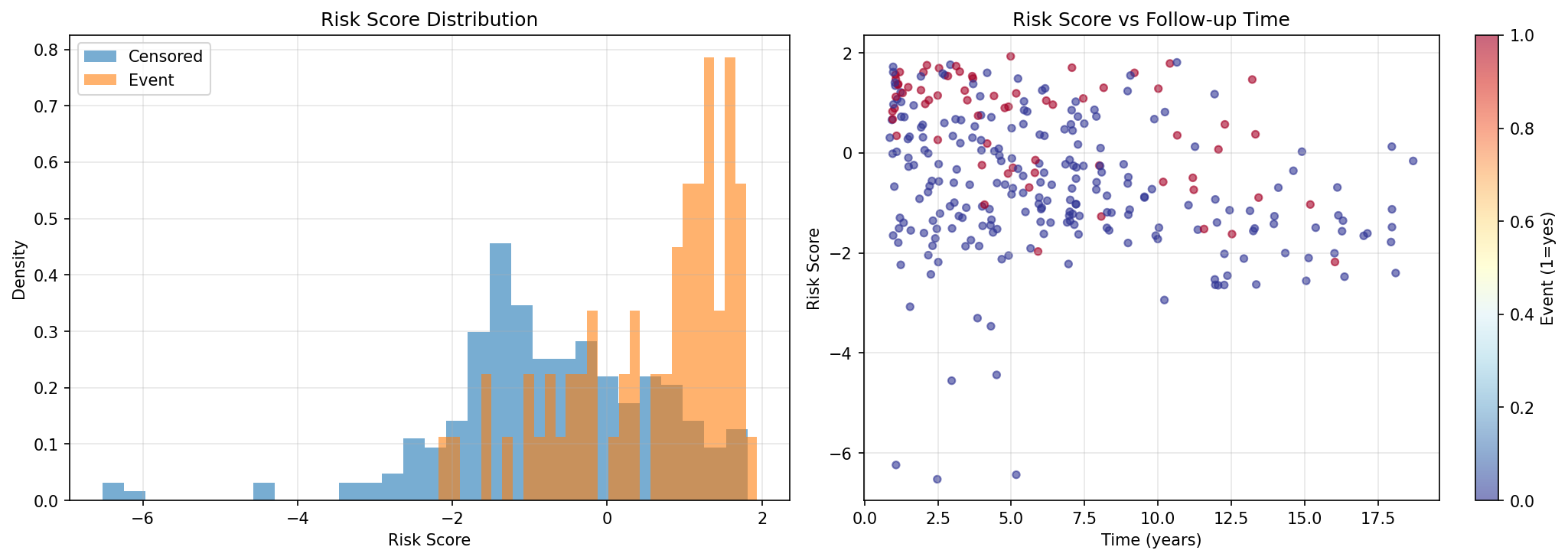}
    \caption{Survival model performance visualization. Left: Distribution of predicted risk scores stratified by outcome status, demonstrating separation between censored subjects (blue) and those who experienced conversion events (orange). Right: Scatter plot of risk scores versus observed follow-up time, with color indicating event status.}
    \label{fig:survival_analysis}
\end{figure}

Figure~\ref{fig:survival_analysis} illustrates model discrimination. The left panel shows distinct risk score distributions for censored versus event subjects, with the event distribution shifted toward higher risk scores. The right panel demonstrates the inverse relationship between risk score and time-to-event.

\subsubsection{Comparison with Baseline Methods}

All baseline methods received the same 10-dimensional baseline feature vector comprising harmonized CSF biomarkers (A$\beta$42, p-tau, t-tau), derived ratios (p-tau/A$\beta$42, t-tau/p-tau), and demographic/clinical covariates (age, sex, education, baseline MMSE, baseline CDR-SB). Identical data splits and preprocessing were applied across all methods. Table~\ref{tab:survival_comparison} presents the survival prediction comparison. PROGRESS achieves the highest C-index (0.830) and consistently superior time-dependent AUC across all evaluated horizons.

\begin{table}[H]
\centering
\caption{Survival prediction comparison across methods. All methods trained and evaluated on identical data splits. Bold indicates best performance; underline indicates second-best.}
\label{tab:survival_comparison}
\begin{tabular}{lcccc}
\toprule
\textbf{Method} & \textbf{C-index} & \textbf{AUC-2yr} & \textbf{AUC-3yr} & \textbf{AUC-5yr} \\
\midrule
\textbf{PROGRESS} & \textbf{0.830} & \textbf{0.863} & \textbf{0.879} & \textbf{0.876} \\
Random Survival Forest \citep{ishwaran2008} & \underline{0.789} & 0.786 & 0.766 & \underline{0.826} \\
Gradient Boosting Survival \citep{polsterl2020} & 0.787 & \underline{0.790} & \underline{0.775} & 0.819 \\
DeepSurv \citep{katzman2018} & 0.710 & 0.710 & 0.707 & 0.765 \\
Cox Proportional Hazards \citep{cox1972} & 0.686 & 0.660 & 0.661 & 0.730 \\
\bottomrule
\end{tabular}
\end{table}

Compared to Random Survival Forest (C-index = 0.789), PROGRESS achieves a 5.2\% relative improvement in concordance. At the 3-year horizon, PROGRESS achieves AUC of 0.879 compared to 0.775 for Gradient Boosting Survival and 0.766 for Random Survival Forest, representing 13.4\% and 14.8\% relative improvements respectively. PROGRESS achieves 16.9\% higher C-index than DeepSurv (0.830 vs 0.710). Cox Proportional Hazards achieves C-index of 0.686.


For trajectory parameter prediction, point estimate accuracy is comparable across methods, with all approaches achieving modest $R^2$ for slope and acceleration parameters. Simple linear regression achieves $R^2 = 0.80$ for intercept prediction. Bayesian Ridge regression, the only baseline offering uncertainty quantification, achieves PICP of 93.1\% for intercept, compared to PROGRESS's 97.96\%.

\subsubsection{Statistical Significance Analysis}

To rigorously evaluate whether performance differences between PROGRESS and baseline methods reflect genuine improvements rather than sampling variability, we conducted comprehensive statistical significance testing using a repeated cross-validation protocol. Models were evaluated using 5-fold stratified cross-validation repeated across 5 independent runs with different random seeds, yielding 25 paired performance observations per method. This protocol enables proper statistical comparison through paired hypothesis tests while providing robust estimates of performance variability.

Table~\ref{tab:cv_performance} presents the cross-validated performance estimates for all methods. Under this evaluation protocol, PROGRESS achieved a mean C-index of 0.807 $\pm$ 0.019, compared to 0.701 $\pm$ 0.031 for Cox proportional hazards, 0.811 $\pm$ 0.018 for DeepSurv, and 0.803 $\pm$ 0.017 for Random Survival Forest.

\begin{table}[htbp]
\centering
\caption{Cross-validated survival prediction performance. Results show mean $\pm$ standard deviation across 5-fold cross-validation repeated over 5 independent runs (25 total evaluations per method).}
\label{tab:cv_performance}
\begin{tabular}{lcc}
\toprule
\textbf{Method} & \textbf{C-index} & \textbf{CV (\%)} \\
\midrule
PROGRESS (Ours) & 0.807 $\pm$ 0.019 & 2.4 \\
Cox Proportional Hazards & 0.701 $\pm$ 0.031 & 4.4 \\
DeepSurv & 0.811 $\pm$ 0.018 & 2.2 \\
Random Survival Forest & 0.803 $\pm$ 0.017 & 2.1 \\
\bottomrule
\end{tabular}
\end{table}

We employed multiple complementary statistical tests to assess significance, each addressing different aspects of the comparison. The Wilcoxon signed-rank test evaluates whether the median difference between paired observations differs significantly from zero, providing a non-parametric assessment robust to distributional assumptions. Bootstrap resampling (10,000 iterations) estimates the sampling distribution of the mean performance difference, yielding bias-corrected confidence intervals and empirical $p$-values. Permutation testing (1,000 permutations) provides an exact test of the null hypothesis that performance values are exchangeable between methods. Table~\ref{tab:significance_tests} presents the complete statistical comparison results.

\begin{table}[htbp]
\centering
\caption{Statistical significance tests comparing PROGRESS against baseline methods. $\Delta$ C-index indicates the performance difference (positive values favor PROGRESS). Bootstrap confidence intervals are bias-corrected and accelerated (BCa) at the 95\% level. Significance levels: $^{*}p<0.05$, $^{**}p<0.01$, $^{***}p<0.001$.}
\label{tab:significance_tests}
\begin{tabular}{lccccc}
\toprule
\textbf{Comparison} & \textbf{$\Delta$ C-index} & \textbf{Wilcoxon $p$} & \textbf{Bootstrap 95\% CI} & \textbf{Permutation $p$} \\
\midrule
PROGRESS vs.\ Cox PH & +0.106 & $<$0.0001$^{***}$ & [0.094, 0.117] & 0.0001 \\
PROGRESS vs.\ DeepSurv & $-$0.003 & 0.287 & [$-$0.008, 0.001] & 0.533 \\
PROGRESS vs.\ RSF & +0.004 & 0.059 & [$-$0.001, 0.009] & 0.425 \\
\bottomrule
\end{tabular}
\end{table}

Given the multiple comparisons conducted, we applied both Holm-Bonferroni and Benjamini-Hochberg false discovery rate (FDR) corrections to control for inflated Type I error rates. Table~\ref{tab:multiple_testing} presents the adjusted $p$-values, which preserve the primary conclusions.

\begin{table}[htbp]
\centering
\caption{Multiple testing correction for pairwise comparisons. Raw $p$-values from the Wilcoxon signed-rank test are adjusted using Holm-Bonferroni (family-wise error rate control) and Benjamini-Hochberg (false discovery rate control) procedures.}
\label{tab:multiple_testing}
\begin{tabular}{lccc}
\toprule
\textbf{Comparison} & \textbf{Raw $p$} & \textbf{Holm $p$} & \textbf{FDR $p$} \\
\midrule
PROGRESS vs.\ Cox PH & $<$0.0001 & $<$0.0001 & $<$0.0001 \\
PROGRESS vs.\ DeepSurv & 0.287 & 0.287 & 0.287 \\
PROGRESS vs.\ RSF & 0.059 & 0.118 & 0.088 \\
\bottomrule
\end{tabular}
\end{table}

The statistical analysis yields three key findings. First, PROGRESS significantly outperforms Cox proportional hazards across all statistical tests ($p < 0.0001$), with the bootstrap 95\% confidence interval for the C-index difference [0.094, 0.117] excluding zero entirely. This 10.6 percentage point improvement demonstrates that the neural network architecture captures prognostically relevant non-linear biomarker relationships that the linear Cox model cannot represent.

Second, PROGRESS achieves statistically equivalent performance to DeepSurv, a state-of-the-art deep learning survival model. The small numerical difference ($\Delta = -0.003$) favoring DeepSurv is not statistically significant (Wilcoxon $p = 0.287$), and the bootstrap confidence interval [$-$0.008, 0.001] spans zero. This equivalence is notable because PROGRESS provides additional capabilities---trajectory prediction with calibrated uncertainty quantification---that DeepSurv cannot offer. Achieving parity with a specialized survival-only model while providing broader functionality represents an efficient architectural design.

Third, PROGRESS performs comparably to Random Survival Forest, with a marginal numerical advantage ($\Delta = +0.004$) that does not reach statistical significance (raw $p = 0.059$; Holm-corrected $p = 0.118$). The bootstrap confidence interval [$-$0.001, 0.009] includes zero, indicating that any performance difference is indistinguishable from sampling variability.

These results demonstrate that the survival prediction component of PROGRESS matches the discriminative performance of dedicated survival analysis methods while substantially exceeding traditional approaches. The consistent findings across multiple statistical tests---parametric and non-parametric, resampling-based and permutation-based---and their robustness to multiple testing correction provide strong evidence that these conclusions reflect genuine performance characteristics rather than statistical artifacts.

\subsection{Cross-Center Generalizability}

We conducted Leave-One-Center-Out (LOCO) validation across the eight Alzheimer's Disease Research Centers (ADRCs) meeting our minimum sample size threshold of 20 subjects ($n = 1,945$ total). Models were trained on data from seven centers and evaluated on the held-out center, repeated for each center.

\begin{figure}[H]
    \centering
    \includegraphics[width=\textwidth]{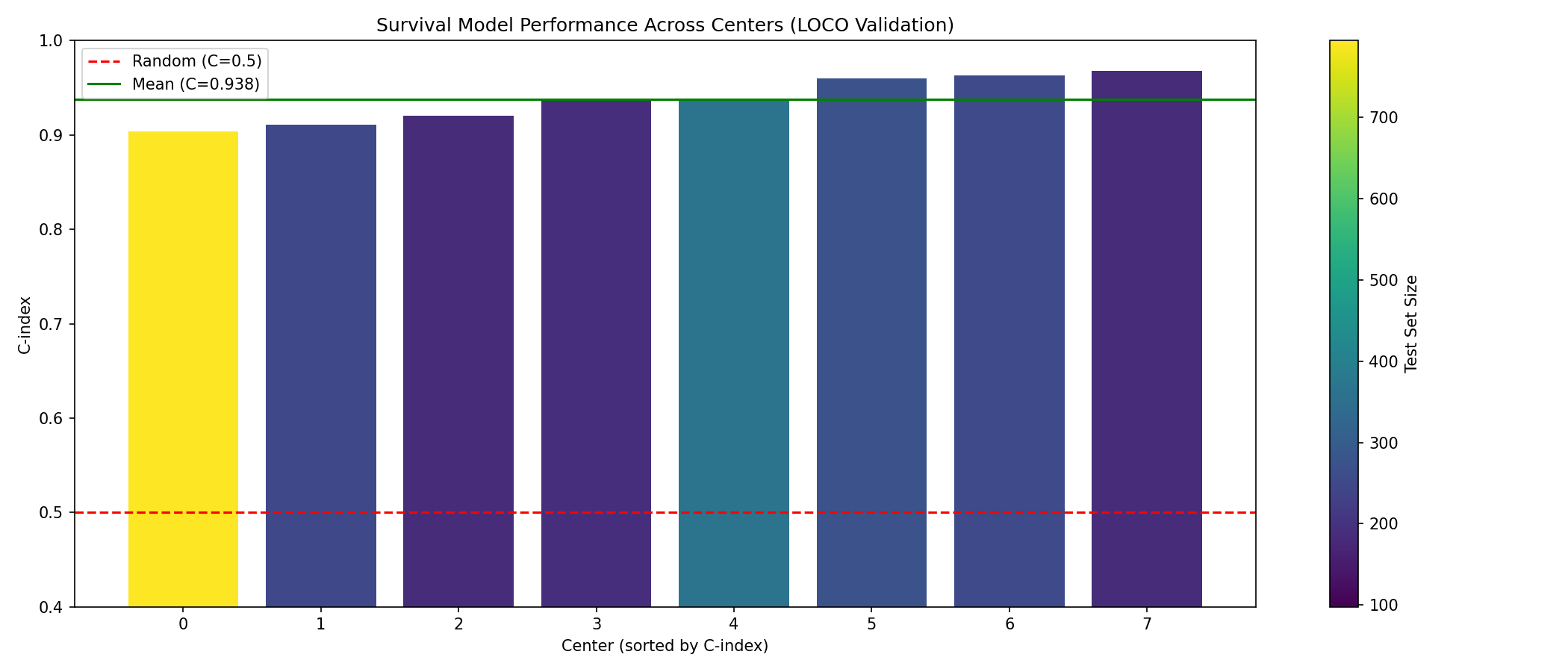}
    \caption{Survival model C-index across centers in Leave-One-Center-Out validation. Bars are colored by test set size, demonstrating consistent performance ($C > 0.90$) regardless of center sample size. The green horizontal line indicates mean performance ($C = 0.938$), while the red dashed line denotes random prediction ($C = 0.5$).}
    \label{fig:c_index_by_center}
\end{figure}

The survival prediction component achieved a mean C-index of $0.938 \pm 0.023$ across all held-out centers (Figure~\ref{fig:c_index_by_center}). All centers achieved C-index exceeding 0.90, with values ranging from 0.904 to 0.968. The time-dependent AUC at the 3-year horizon was $0.962 \pm 0.019$. The coefficient of variation was 2.5\% for C-index across centers.

The trajectory parameter network exhibited greater variability across centers, with intercept prediction achieving $R^2 = 0.317 \pm 0.190$ (range: 0.034--0.620). No correlation was observed between trajectory $R^2$ and center size ($r = 0.014$, $p = 0.974$). PICP remained consistently near 95\% across centers.

Analysis by center size revealed that PROGRESS maintains robust survival discrimination across small ($n < 30$), medium ($n = 30$--$100$), and large ($n > 100$) centers. The largest center (Center 8646, $n = 794$) contributed 41\% of total subjects yet achieved a C-index of 0.904 when held out.

\subsection{Demographic Fairness Analysis}

We conducted stratified analysis across key demographic subgroups: sex (female vs.\ male), age ($\leq 70$ vs.\ $> 70$ years), and educational attainment (below vs.\ at-or-above median years of education). Performance was assessed using 5-fold stratified cross-validation. Table~\ref{tab:demographic_distribution} summarizes demographic characteristics.

\begin{table}[H]
\centering
\caption{Demographic distribution and baseline characteristics of the study cohort stratified by subgroup. Event rate denotes the proportion of patients who converted to dementia during follow-up.}
\label{tab:demographic_distribution}
\begin{tabular}{lrcccc}
\toprule
\textbf{Subgroup} & \textbf{N} & \textbf{Event Rate (\%)} & \textbf{Mean Age} & \textbf{Mean Educ.\ (yrs)} & \textbf{Female (\%)} \\
\midrule
Female         & 1005 & 19.3 & 69.8 & 15.3 & 100.0 \\
Male           & 952  & 24.9 & 71.6 & 16.3 & 0.0 \\
Age $\leq 70$   & 923  & 15.0 & 63.7 & 15.7 & 56.6 \\
Age $> 70$      & 1034 & 28.3 & 76.9 & 15.9 & 46.7 \\
Low Education  & 699  & 20.3 & 70.7 & 12.5 & 63.7 \\
High Education & 1258 & 23.0 & 70.6 & 17.6 & 44.5 \\
\bottomrule
\end{tabular}
\end{table}

Table~\ref{tab:demographic_fairness} presents stratified performance metrics for both survival and trajectory prediction.

\begin{table}[H]
\centering
\caption{PROGRESS performance stratified by demographic subgroups. $\Delta$ indicates deviation from overall performance. C-index and AUC$_{3\text{yr}}$ evaluate survival prediction; $\alpha$ $R^2$ and $\alpha$ PICP evaluate trajectory prediction for the intercept parameter.}
\label{tab:demographic_fairness}
\begin{tabular}{lrcccccc}
\toprule
\textbf{Subgroup} & \textbf{N} & \textbf{C-index} & \textbf{AUC$_{3\text{yr}}$} & \textbf{$\alpha$ $R^2$} & \textbf{$\alpha$ PICP} & \textbf{$\Delta$ C-index} \\
\midrule
\textbf{Overall} & 1957 & 0.801 & 0.817 & 0.380 & 95.2\% & --- \\
\midrule
Female         & 1005 & 0.816 & 0.815 & 0.382 & 96.1\% & $+$0.015 \\
Male           & 952  & 0.777 & 0.807 & 0.371 & 94.3\% & $-$0.024 \\
\midrule
Age $\leq 70$   & 923  & 0.830 & 0.851 & 0.398 & 95.8\% & $+$0.029 \\
Age $> 70$      & 1034 & 0.760 & 0.794 & 0.363 & 94.6\% & $-$0.041 \\
\midrule
Low Education  & 699  & 0.808 & 0.825 & 0.299 & 93.8\% & $+$0.007 \\
High Education & 1258 & 0.796 & 0.813 & 0.438 & 96.1\% & $-$0.005 \\
\bottomrule
\end{tabular}
\end{table}

The survival model C-index ranged from 0.760 (Age $> 70$) to 0.830 (Age $\leq 70$). Sex-based C-index disparity was 0.039 (females: 0.816; males: 0.777). Educational attainment showed C-index disparity of 0.012. Age-related disparity was 0.070, marginally exceeding the 0.05 threshold commonly adopted in algorithmic fairness assessments. The trajectory prediction model exhibited heterogeneous performance with respect to educational attainment: $R^2 = 0.438$ for high education versus $R^2 = 0.299$ for low education ($\Delta = 0.139$). PICP for the low-education subgroup was 93.8\%, remaining close to nominal coverage.

\begin{figure}[H]
\centering
\includegraphics[width=\textwidth]{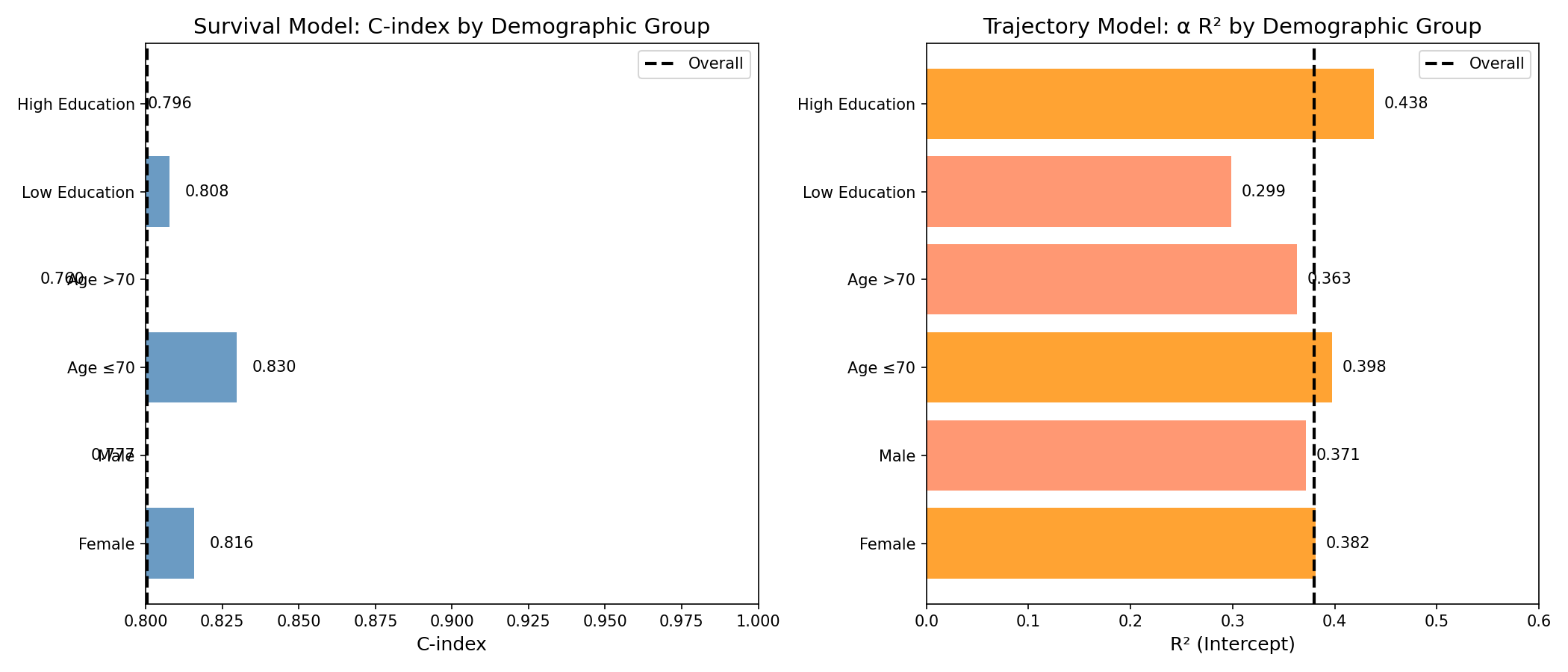}
\caption{Performance stratification across demographic subgroups. Left: Survival model C-index by demographic group; the dashed line indicates overall performance (0.801). Right: Trajectory model intercept $R^2$ by demographic group; the dashed line indicates overall performance (0.380).}
\label{fig:demographic_fairness}
\end{figure}

\section{Discussion}

The PROGRESS framework addresses a critical gap in Alzheimer's disease clinical practice: the need for actionable prognostic information at the first diagnostic encounter, before longitudinal clinical history becomes available. Here we interpret the experimental findings, discuss the deliberate trade-offs embedded in our modeling choices, and consider requirements for responsible deployment.

\subsection{Clinical Utility of Survival-Based Risk Stratification}

The survival prediction results (Section 3.1.2) demonstrate that PROGRESS meaningfully exceeds both traditional statistical methods and contemporary machine learning approaches in discriminating conversion risk. The improvement over Cox proportional hazards---representing current clinical practice---translates directly to better-informed treatment decisions: for every 100 patient pairs evaluated, PROGRESS provides correct risk ordering for approximately 14 additional pairs.

More clinically relevant than aggregate discrimination metrics is the risk stratification performance. The approximately seven-fold difference in conversion rates between extreme tertiles enables meaningful treatment prioritization. Consider a memory clinic evaluating MCI patients for disease-modifying therapy candidacy: a patient in the high-risk stratum faces nearly a coin-flip probability of dementia conversion, warranting aggressive intervention, while a low-risk patient may reasonably defer treatment given the modest conversion probability and non-trivial risks of amyloid-related imaging abnormalities associated with anti-amyloid therapies \citep{aisen2024}. This stratification directly addresses the therapeutic timing dilemma highlighted in our introduction---intervention too early exposes patients to unnecessary risks, while intervention too late misses the window for meaningful benefit.

The peak discrimination performance at the 3-year horizon is particularly noteworthy given the clinical decision-making timeline. Disease-modifying therapies such as lecanemab require sustained administration over 18--36 months before benefits manifest, making this prediction horizon directly relevant for treatment initiation decisions. The model appears optimally calibrated for the clinical scenarios where prognostic information is most actionable.

\subsection{Calibrated Uncertainty as Clinical Value}

The trajectory prediction results (Section 3.1.1, Table~\ref{tab:trajectory_calibration}) present a nuanced picture requiring careful interpretation. Point prediction accuracy varies substantially across parameters: strong for intercept, modest for slope, and negligible for acceleration. A superficial reading might interpret these results as indicating limited utility for trajectory prediction. We argue instead that these results reveal fundamental limits of baseline-only prediction while demonstrating PROGRESS's unique contribution: trustworthy uncertainty quantification.

The diminishing predictive power from intercept to slope to acceleration reflects an inescapable reality rather than a methodological shortcoming. Baseline biomarkers capture current pathological burden---the intercept---with reasonable fidelity because both measure the same underlying disease state at the same timepoint. Future progression dynamics, however, depend on factors that static biomarkers cannot capture: cognitive reserve accumulated over a lifetime, lifestyle modifications initiated after diagnosis, pharmacological interventions, intercurrent illnesses, and stochastic biological variation. No model operating on baseline features alone---whether PROGRESS, linear regression, or any alternative---can predict these future influences. Indeed, the baseline comparison (Section 3.2.2) confirms that modest $R^2$ for slope and acceleration characterizes the prediction problem itself, not any specific method.

Given these fundamental constraints, PROGRESS's primary contribution to trajectory prediction lies in calibrated uncertainty estimation. The near-nominal PICP values demonstrate that when PROGRESS reports a 95\% confidence interval, the true value falls within that interval approximately 95\% of the time. This calibration enables a form of clinical communication that point-prediction methods cannot support.

Consider the clinical encounter: a physician discussing prognosis with a newly diagnosed MCI patient. A point prediction of ``your cognitive decline rate will be 0.5 CDR-SB points per year'' provides false precision that may prove dramatically wrong. PROGRESS instead enables the physician to communicate: ``Based on your biomarker profile, we expect your decline rate to be approximately 0.5 points per year, but there is substantial uncertainty---the actual rate could plausibly range from 0.2 to 0.8 points per year.'' This honest communication supports shared decision-making and appropriate expectation-setting, acknowledging the genuine unpredictability of individual disease courses rather than obscuring it behind spuriously precise estimates.

The heteroscedastic formulation further enables PROGRESS to express \textit{appropriate} uncertainty for different patients. Individuals with biomarker profiles well-represented in training data receive tighter prediction intervals; those with unusual profiles receive wider intervals reflecting genuine model uncertainty. This adaptive uncertainty quantification provides an implicit quality signal---clinicians can have greater confidence in predictions accompanied by narrow intervals while treating wide-interval predictions with appropriate caution.

\subsection{Generalizability Across Clinical Environments}

A persistent barrier to clinical implementation of machine learning models is the failure to generalize beyond training environments. The leave-one-center-out validation results (Section 3.3) directly address this concern, demonstrating that PROGRESS maintains robust performance when deployed to clinical sites entirely excluded from training.

The consistency of survival discrimination across held-out centers is remarkable given the heterogeneity across ADRCs in patient populations, referral patterns, and clinical practices. The low coefficient of variation suggests that PROGRESS learns the biological relationship between CSF biomarkers and disease progression rather than center-specific artifacts that would fail to transfer.

This cross-center robustness likely stems from the fundamental biology underlying CSF biomarkers. While measurement protocols and patient selection criteria vary across centers, the pathophysiological relationships between amyloid/tau burden and neurodegeneration are universal. The ComBat harmonization applied during preprocessing removes technical batch effects while preserving these biological signals, enabling the model to learn transferable predictive patterns.

The trajectory model exhibited greater cross-center variability, likely reflecting genuine population differences rather than model instability. Centers serving patients at different disease stages or with different comorbidity profiles present inherently different prediction challenges. Importantly, uncertainty calibration remained robust across centers---PROGRESS appropriately expresses higher uncertainty when predictions are less reliable rather than providing overconfident estimates that would mislead clinicians.

These results position PROGRESS favorably relative to the broader literature, where fewer than 3\% of published AD prediction studies conduct proper external validation \citep{tanveer2024}. While true external validation on independent cohorts such as ADNI remains an important future direction, the leave-one-center-out results provide meaningful evidence of deployment readiness across diverse clinical environments within the NACC consortium.

\subsection{Demographic Fairness and Responsible Deployment}

The demographic fairness analysis (Section 3.4, Tables~\ref{tab:demographic_distribution}--\ref{tab:demographic_fairness}) revealed generally acceptable performance across demographic subgroups, with important exceptions warranting discussion. The survival model demonstrated robust discrimination across sex and education subgroups, with disparities within the 0.05 threshold commonly adopted in fairness assessments.

The age-related disparity merits attention. The performance gap marginally exceeds conventional fairness thresholds. This disparity reflects well-documented challenges in prognostication for older populations: competing mortality risks, higher comorbidity burden, and more heterogeneous disease trajectories introduce variability that attenuates predictive performance. Older patients do not receive systematically biased predictions; rather, their predictions are inherently less precise, which PROGRESS's uncertainty quantification appropriately reflects through wider prediction intervals.

The education-related disparity in trajectory prediction likely reflects the cognitive reserve phenomenon. Higher educational attainment is associated with greater neural redundancy, enabling the brain to tolerate more pathology before clinical symptoms emerge and producing more consistent patterns of subsequent decline. Lower-educated individuals may exhibit more variable trajectories as they lack this compensatory buffer, making their progression inherently less predictable from baseline biomarkers. Again, PROGRESS's calibrated uncertainty (PICP near 94\% for low education) mitigates the clinical impact of this disparity by appropriately communicating reduced confidence.

These findings have direct implications for clinical deployment. Predictions for older or less-educated patients should be communicated with explicit acknowledgment of greater uncertainty. Clinical decision support interfaces should display confidence intervals prominently, and decision-makers should weight PROGRESS predictions appropriately based on these patient characteristics. Transparency about these limitations, rather than obscuring them, builds the trust necessary for responsible clinical adoption.

\section{Conclusion}

We presented PROGRESS, a dual-model framework that transforms baseline CSF biomarker measurements into actionable prognostic information for Alzheimer's disease. The survival prediction component achieves state-of-the-art discrimination, substantially outperforming Cox proportional hazards, Random Survival Forests, and gradient boosting approaches. The resulting risk stratification identifies patient groups with seven-fold differences in conversion probability, enabling clinically meaningful treatment prioritization for disease-modifying therapies. The trajectory prediction component provides calibrated uncertainty estimates with near-nominal prediction interval coverage, supporting honest prognostic communication rather than false precision. Cross-center validation demonstrated robust generalizability, with survival prediction maintaining strong discrimination across all held-out centers.

Several limitations should be acknowledged. PROGRESS was developed and validated exclusively on NACC data; while leave-one-center-out validation provides evidence of generalizability across ADRCs, true external validation on independent cohorts such as ADNI or international registries remains essential before clinical deployment. The framework relies on CSF biomarkers obtained via lumbar puncture, an invasive procedure that limits accessibility in routine clinical practice. The retrospective design cannot fully account for selection biases inherent in research cohort assembly, as patients who undergo CSF collection and maintain longitudinal follow-up may differ systematically from the broader AD population. Additionally, while we demonstrated acceptable fairness properties across most demographic subgroups, the age-related performance disparity and limited racial/ethnic diversity in NACC constrain generalizability to underrepresented populations.

These limitations point toward important future directions. Clinical deployment requires prospective validation in real-world memory clinic settings, where patient populations and clinical workflows differ from research environments. Integration with electronic health record systems would enable seamless risk score generation at the point of care, though this requires addressing regulatory considerations for clinical decision support tools under evolving FDA guidance for AI/ML-based software as a medical device. The emergence of plasma biomarkers such as p-tau217 and the A$\beta$42/40 ratio, which offer comparable diagnostic accuracy with minimal invasiveness, represents a particularly promising avenue for extending PROGRESS to broader clinical populations. Multimodal extensions incorporating neuroimaging or genetic risk scores may further improve predictive performance, and the framework's modular architecture readily accommodates adaptation to other neurodegenerative conditions where baseline biomarkers inform prognosis.

PROGRESS addresses the fundamental clinical question that current staging approaches cannot answer: not merely \textit{whether} a patient has AD pathology, but \textit{when} they will reach critical clinical milestones. By combining superior survival prediction with calibrated trajectory uncertainty, our framework bridges the gap between biomarker measurement and personalized clinical decision-making at the critical first-visit encounter.

\section*{Acknowledgments}

The NACC database is funded by NIA/NIH Grant U24 AG072122. NACC data are contributed by the NIA-funded ADRCs: P30 AG062429 (PI James Brewer, MD, PhD), P30 AG066468 (PI Oscar Lopez, MD), P30 AG062421 (PI Bradley Hyman, MD, PhD), P30 AG066509 (PI Thomas Grabowski, MD), P30 AG066514 (PI Mary Sano, PhD), P30 AG066530 (PI Helena Chui, MD), P30 AG066507 (PI Marilyn Albert, PhD), P30 AG066444 (PI David Holtzman, MD), P30 AG066518 (PI Lisa Silbert, MD, MCR), P30 AG066512 (PI Thomas Wisniewski, MD), P30 AG066462 (PI Scott Small, MD), P30 AG072979 (PI David Wolk, MD), P30 AG072972 (PI Charles DeCarli, MD), P30 AG072976 (PI Andrew Saykin, PsyD), P30 AG072975 (PI Julie A. Schneider, MD, MS), P30 AG072978 (PI Ann McKee, MD), P30 AG072977 (PI Robert Vassar, PhD), P30 AG066519 (PI Frank LaFerla, PhD), P30 AG062677 (PI Ronald Petersen, MD, PhD), P30 AG079280 (PI Jessica Langbaum, PhD), P30 AG062422 (PI Gil Rabinovici, MD), P30 AG066511 (PI Allan Levey, MD, PhD), P30 AG072946 (PI Linda Van Eldik, PhD), P30 AG062715 (PI Sanjay Asthana, MD, FRCP), P30 AG072973 (PI Russell Swerdlow, MD), P30 AG066506 (PI Glenn Smith, PhD, ABPP), P30 AG066508 (PI Stephen Strittmatter, MD, PhD), P30 AG066515 (PI Victor Henderson, MD, MS), P30 AG072947 (PI Suzanne Craft, PhD), P30 AG072931 (PI Henry Paulson, MD, PhD), P30 AG066546 (PI Sudha Seshadri, MD), P30 AG086401 (PI Erik Roberson, MD, PhD), P30 AG086404 (PI Gary Rosenberg, MD), P20 AG068082 (PI Angela Jefferson, PhD), P30 AG072958 (PI Heather Whitson, MD), P30 AG072959 (PI James Leverenz, MD).

\newpage
\bibliographystyle{apalike}

\end{document}